%% file: main.tex
\documentclass[11pt]{article}
\usepackage[margin=1in]{geometry}
\usepackage[T1]{fontenc}
\usepackage[utf8]{inputenc}

\usepackage{microtype}
\usepackage{graphicx}
\usepackage{subcaption}
\usepackage{booktabs}
\usepackage{amsmath,amssymb,mathtools}
\usepackage{siunitx}
\usepackage{enumitem}
\usepackage{xcolor}
\usepackage{soul,xcolor}
\sethlcolor{red!20}
\usepackage{hyperref}
\usepackage[capitalize,noabbrev]{cleveref}
\usepackage[numbers,sort&compress]{natbib}
\usepackage{algorithm}
\usepackage{algpseudocode}
\usepackage{multirow}
\usepackage{colortbl}
\usepackage{pifont}
\usepackage{url}
\usepackage{float}

\hypersetup{
  colorlinks=true,
  linkcolor=black,
  citecolor=blue!60!black,
  urlcolor=blue!60!black
}

\newcommand{\method}{RAGSmith} 
\newcommand{\fullname}{A Framework for Finding the Optimal Composition of Retrieval-Augmented Generation Methods Across Datasets}

\newcommand{\bm}[1]{\boldsymbol{#1}}

\title{\method: \fullname\\}

\newif\ifanonymous
\anonymousfalse

\ifanonymous
\author{Anonymous Authors}
\date{}
\else
\author{
  Muhammed Yusuf Kartal$^{1}$\;\;
  Suha Kagan Kose$^{2}$\;\;
  Korhan Sevin\c{c}$^{1}$\;\;
  Burak Aktas$^{2}$\\
  $^{1}$TOBB University of Economics and Technology\quad $^{2}$Roketsan Inc.\quad\\
  \texttt{m.kartal@etu.edu.tr}\;\; \texttt{kagan.kose@roketsan.com.tr}\\
  \texttt{ksevinc@etu.edu.tr}\;\; \texttt{burak.aktas@roketsan.com.tr}
}
\date{\today}
\fi

\begin{document}
\maketitle

\begin{abstract}
Retrieval-Augmented Generation (RAG) quality depends on many interacting choices across retrieval, ranking, augmentation, prompting, and generation, so optimizing modules in isolation is brittle. We introduce RAGSmith, a modular framework that treats RAG design as an end-to-end architecture search over nine technique families and 46{,}080 feasible pipeline configurations. A genetic search optimizes a scalar objective that jointly aggregates retrieval metrics (recall@k, mAP, nDCG, MRR) and generation metrics (LLM-Judge and semantic similarity). We evaluate on six Wikipedia-derived domains (Mathematics, Law, Finance, Medicine, Defense Industry, Computer Science), each with 100 questions spanning factual, interpretation, and long-answer types. RAGSmith finds configurations that consistently outperform naive RAG baseline by +3.8\% on average (range +1.2\% to +6.9\% across domains), with gains up to +12.5\% in retrieval and +7.5\% in generation. The search typically explores $\approx 0.2\%$ of the space ($\sim 100$ candidates) and discovers a robust backbone---vector retrieval plus post-generation reflection/revision---augmented by domain-dependent choices in expansion, reranking, augmentation, and prompt reordering; passage compression is never selected. Improvement magnitude correlates with question type, with larger gains on factual/long-answer mixes than interpretation-heavy sets. These results provide practical, domain-aware guidance for assembling effective RAG systems and demonstrate the utility of evolutionary search for full-pipeline optimization.
\end{abstract}

\input{sections/01-introduction}
\input{sections/02-related_work}
\input{sections/03-methods}

\input{sections/04-experiment}
\input{sections/05-results_and_discussion}
\input{sections/06-conclusion}
\input{sections/07-acknowledgement}

\bibliographystyle{ieeetr}
\bibliography{references}
\end{document}

%% file: sections/01-introduction.tex
\section{Introduction}

Large Language Models (LLMs) have demonstrated remarkable capabilities in natural language understanding and generation, yet they face persistent challenges with factual accuracy, knowledge currency, and domain-specific expertise. Retrieval-Augmented Generation (RAG)~\cite{lewis2020retrieval} has emerged as a powerful paradigm to address these limitations by augmenting LLM generation with relevant information retrieved from external knowledge sources. By grounding responses in retrieved evidence, RAG systems can provide more accurate, up-to-date, and verifiable answers while mitigating hallucination—a critical concern in deploying LLMs for knowledge-intensive tasks.

The basic RAG architecture comprises three stages: retrieval of relevant documents from a knowledge base, augmentation of the query context with retrieved passages, and generation of responses conditioned on both the original query and retrieved information. However, this simple pipeline has evolved substantially. Advanced RAG techniques now encompass sophisticated components including query expansion~\cite{wang2023query}, multi-stage reranking~\cite{lin2024reranking}, contextual embeddings~\cite{li2024contextual}, and iterative refinement strategies~\cite{shinn2024reflexion}. These techniques aim to enhance retrieval precision, improve context utilization, and refine generated responses, collectively pushing RAG performance beyond naive implementations.

Despite this proliferation of advanced RAG components, a fundamental challenge remains: optimizing RAG pipelines has largely followed a greedy, per-module approach. Existing work typically evaluates retrieval methods, reranking strategies, and generation techniques independently, then combines the individually best-performing components into a final system. For instance, researchers might compare embedding models in isolation, select the top performer, then separately evaluate rerankers to find the best option~\cite{gao2023survey}. While this modular evaluation provides insights into individual technique effectiveness, it fundamentally ignores potential synergies and conflicts between components. A retrieval method that excels with one reranker may underperform with another; an augmentation strategy beneficial for one generation approach may hinder another. The assumption that locally optimal choices yield globally optimal configurations is untenable when components interact in complex, non-linear ways.

This limitation is compounded by the domain-specificity of RAG optimization. Optimal configurations vary substantially across knowledge domains due to differences in content structure, terminology consistency, and question types. A RAG pipeline tuned for medical literature—with its standardized terminology and hierarchical organization—may perform poorly on legal documents with their variable section lengths and dense cross-references. Similarly, the effectiveness of specific techniques depends on dataset characteristics: query expansion may benefit domains with lexical variation, while multi-stage reranking proves essential for high-density knowledge bases. Existing optimization approaches, by treating domains uniformly and optimizing components independently, fail to capture these domain-specific patterns and inter-component dependencies.

In this work, we present \method, a framework for holistic RAG pipeline optimization through evolutionary search. Our key insight is that RAG configuration should be treated as a complete-pipeline optimization problem rather than independent module selection. We introduce a modular pipeline architecture encompassing nine stages—from pre-embedding enrichment through post-generation refinement—and implement 46,080 possible configurations. Using genetic algorithms, we evolve complete pipeline configurations, evaluating fitness holistically and allowing modules that may be suboptimal in isolation to emerge when they enhance overall system performance.

We apply \method~to six domain-specific datasets derived from Wikipedia articles (Mathematics, Law, Finance, Medicine, Defense Industry, and Computer Science), each containing 100 questions across three categories: \textit{factual}, \textit{interpretation}, and \textit{long-answer}. Through comprehensive evaluation comparing optimized configurations against naive RAG baseline, we demonstrate consistent improvements averaging +3.8\% overall performance (ranging from +1.2\% to +6.9\% across domains), with retrieval improvements up to +12.5\% and generation improvements up to +7.5\%.

Our contributions are as follows:

\begin{itemize}
\item \textbf{Holistic Pipeline Optimization}: We introduce an evolutionary approach that optimizes complete RAG configurations rather than independently selecting per-module "best" components, accounting for inter-component synergies and conflicts that greedy optimization ignore. Moreover, since exhaustive grid search is computationally prohibitive given the substantial runtime of end-to-end RAG pipelines, we adopt a genetic search strategy to efficiently explore the configuration space and rapidly identify high-performing RAG settings.

\item \textbf{Question Type Sensitivity Framework}: We establish that question type distribution significantly influences both optimal configuration and improvement potential, with \textit{long-answer} and \textit{factual}-heavy datasets achieving larger gains (+5--7\%) than \textit{interpretation}-heavy datasets (+1--4\%), revealing fundamental limitations in current RAG techniques' ability to enhance inferential reasoning.

\item \textbf{Domain-Specific Optimization Guidelines}: We identify empirically-grounded patterns mapping dataset characteristics to effective technique combinations—chunk density determines reranker selection, information density uniformity guides augmentation strategy choice, and hierarchical content structure motivates pre-embedding enrichment.

\item \textbf{Robust RAG Backbone}: Across all evaluated datasets, we observe that \texttt{vector\_retrieval} (as the retrieval component) and \texttt{reflection\_revising} (as the post-generation refinement component) consistently yield the strongest performance among the explored RAG techniques. This suggests that these two components constitute a robust, subject-agnostic backbone. In contrast, other modules—such as query expansion, reranking, and prompt maker—exhibit more subject-dependent behavior and operate as adaptive extensions that can be tuned to the specifics of a given task or dataset.
\end{itemize}

The remainder of this paper is organized as follows: Section 2 reviews related work in advanced RAG methods and modular pipeline design. Section 3 describes our modular RAG framework and evolutionary search methodology. Section 4 details experimental setup, datasets, and evaluation metrics. Section 5 presents results and discussion of our experiment. Section 6 concludes with key findings and implications for RAG system design. Our codebase and datasets used in this paper can be found at \url{https://github.com/yAquila/RAGSmith}.

%% file: sections/02-related_work.tex
\section{Related Work}
\subsection{Retrieval-Augmented Generation (RAG) Foundations}
\label{sec:rag_foundations}

Retrieval-Augmented Generation (RAG) refers to a general inference-time paradigm in which a language model is provided with externally retrieved evidence before (or while) generating an answer \cite{lewis2020rag}. The core idea is simple: \emph{retrieve first, then generate}. Instead of relying solely on parametric knowledge stored in model weights, a RAG system queries a corpus, selects relevant passages, and conditions the generator on those passages to produce the final output. This retrieve-then-generate loop has become a standard approach for open-domain question answering, knowledge-intensive reasoning, and adaptation to specialized domains \cite{lewis2020rag,karpukhin2020dpr,izacard2021fid}.

A ``vanilla RAG'' pipeline typically has two main components: (i) a retriever, and (ii) a generator. Beyond this high-level structure, such systems are motivated by three goals: grounding generated text in explicit evidence, improving factuality, and enabling domain adaptation without full model fine-tuning.

\paragraph{Retriever.}
The retriever maps the input query to a small set of relevant text chunks drawn from a large corpus. Early open-domain QA systems often relied on sparse lexical retrievers such as TF--IDF or BM25, sometimes followed by a neural reader \cite{chen2017drqa}. More recent pipelines use dense retrieval, where both queries and passages are embedded into a shared vector space using a dual-encoder model trained to bring matching question--passage pairs closer together \cite{karpukhin2020dpr}. Dense Passage Retrieval (DPR) showed that dense bi-encoder retrievers, trained with in-batch negatives and supervised question--evidence pairs, can outperform sparse retrieval on open-domain QA benchmarks while remaining efficient at scale \cite{karpukhin2020dpr}.

Conditioning generation on retrieved text directly supports factual grounding. Because the retriever surfaces concrete evidence at inference time, the downstream generator can attribute claims to specific passages rather than relying purely on internal memorization \cite{lewis2020rag}. This external grounding is important for transparency and verifiability in knowledge-intensive tasks such as open-domain QA and fact attribution \cite{lewis2020rag,izacard2021fid}.

\paragraph{Generator.}
Given the retrieved passages, a generator (usually a sequence-to-sequence transformer) produces an answer conditioned on both the query and the retrieved evidence \cite{lewis2020rag}. Lewis et al.\ introduced the term ``Retrieval-Augmented Generation'' for a model that jointly learns the retriever and the generator: the retriever proposes candidate passages, and the generator marginalizes over them during training and inference \cite{lewis2020rag}. In this setup, generation is explicitly grounded in retrieved knowledge, which helps prevent hallucinated content and encourages factual attribution to specific passages.

This retrieve-then-generate design also supports domain adaptation without fully fine-tuning large models. A pretrained generator can be reused across domains, while retrieval is redirected to a new in-domain corpus (e.g., internal manuals, scientific articles, legal databases) without retraining the full language model \cite{guu2020realm,lewis2020rag}. Dense retrievers such as DPR can also be adapted or further trained on task-specific relevance signals \cite{karpukhin2020dpr}, and FiD-style generators can scale to larger evidence sets from that new corpus \cite{izacard2021fid}. This offers a practical alternative to full supervised fine-tuning of a large model for every new knowledge domain.

One limitation of early RAG-style models is that they often concatenate only a small number of passages into the generator's input, creating a bottleneck when many relevant contexts exist. Fusion-in-Decoder (FiD) addressed this by encoding each retrieved passage independently and letting the decoder attend across all encoded representations, rather than concatenating them into a single long input \cite{izacard2021fid}. This architecture allows conditioning on dozens or even hundreds of retrieved passages without collapsing them into a single flat context window, and it has become a strong baseline for retrieval-augmented question answering and knowledge-intensive generation \cite{izacard2021fid}.

The pipeline above is the de facto baseline that most modern retrieval-augmented systems inherit \cite{karpukhin2020dpr,lewis2020rag,izacard2021fid}. Our framework assumes this baseline and asks a higher-level question: given a particular dataset or deployment domain, \emph{which concrete retrieval and generation design choices (indexing strategy, retriever type, fusion strategy, number of passages, etc.) actually work best in practice?} The work in this study builds on the RAG foundations summarized here and treats them as the starting point for systematic comparison.

\subsection{Retrieval Methods and Indexing Strategies}
\label{sec:retrieval_methods}

Before generation, there is already a large design space in how we (i) score relevance between a query and a candidate passage, and (ii) store and access those passages efficiently. This section surveys common retrieval paradigms and indexing strategies used in modern RAG systems. We emphasize that there is no single ``best'' retrieval configuration: different domains, query styles, latency constraints, and corpus characteristics favor different design choices. This motivates the need for systems that can automatically discover or select high-performing retrieval configurations for a given setting.

\paragraph{Cross-encoder rerankers.}
Another common pattern is a two-stage pipeline: a fast first-stage retriever (sparse, dense, or hybrid) retrieves the top-$k$ candidates, and then a more expensive cross-encoder reranker re-scores those candidates \cite{nogueira2019passage}. A cross-encoder takes the full query and a single candidate passage as joint input to a transformer, and directly predicts a relevance score. Because it can attend across all query and passage tokens simultaneously, it typically achieves higher precision than dual-encoders. The drawback is cost: cross-encoders cannot be run exhaustively over the entire corpus, so they act only on a shortlist.

Cross-encoder reranking has become a standard recipe in open-domain QA and web search: retrieve $k \in [20,100]$ candidates using BM25 or DPR, then apply a cross-encoder BERT ranker to produce the final ordering \cite{nogueira2019passage}. For RAG, this increases the chance that truly relevant passages appear in the top few contexts fed into the generator, which in turn improves grounding and factual accuracy.

\paragraph{Hybrid sparse + dense retrieval.}
Hybrid retrieval fuses lexical and dense signals, for example by linearly combining BM25 scores with dense similarity scores, or by interleaving candidate sets from both retrievers before reranking \cite{luan2021sparse,karpukhin2020dpr}. The intuition is that sparse retrieval excels at entity-level lookup and robustness to domain-specific terminology, while dense retrieval excels at semantic generalization and paraphrase matching. Empirical work has repeatedly shown that hybrids often outperform either method alone, especially in heterogeneous corpora where some answers are phrased almost exactly like the query and others are paraphrased \cite{luan2021sparse}. 

From a RAG perspective, this matters because hybrid retrieval improves \emph{recall@k}---the probability that at least one truly answer-bearing passage is included in the retrieved set. Higher recall@k directly increases the headroom for the downstream generator.

\subsubsection{Index Structure, Chunk Granularity, and Memory Access}
\label{sec:index_structures}

Choosing \emph{how} to store and retrieve passages at scale is as important as choosing \emph{what} scoring function to use.

\paragraph{Approximate nearest neighbor (ANN) indexes.}
Dense retrieval relies on fast nearest-neighbor search in high-dimensional vector spaces. Exact search over millions of vectors can be prohibitively slow, so most systems use approximate nearest neighbor (ANN) data structures. FAISS popularized GPU-accelerated and CPU-efficient ANN search via product quantization, inverted file (IVF) lists, and hierarchical clustering \cite{johnson2017billion}. Other widely used ANN structures include graph-based methods such as Hierarchical Navigable Small World (HNSW) graphs, which enable logarithmic-time approximate search with strong recall in practice \cite{malkov2018efficient}.

Index choice directly affects latency, memory footprint, and recall. For example, highly compressed product-quantized indexes are memory-efficient but may degrade fine-grained similarity; HNSW-style graphs provide excellent recall but can consume more RAM. In RAG deployments, these trade-offs become part of system design: Do we optimize for sub-50ms retrieval? For minimal GPU usage? For highest recall at any cost?

\paragraph{Single-vector vs.\ multi-vector storage.}
Classical dense retrievers such as DPR or SBERT store one fixed-size embedding per passage \cite{karpukhin2020dpr,reimers2019sentencebert}. This enables simple inner-product search: each passage corresponds to a single vector in the ANN index.

Late-interaction models such as ColBERT instead store \emph{multiple} contextualized token embeddings per passage \cite{khattab2020colbert}. This ``multi-vector'' representation increases index size but supports more expressive matching at retrieval time, because the query can selectively attend to different parts of the passage without having to compress all semantics into a single vector. Many modern multi-vector indexes exploit inverted-list or asymmetric distance computations to keep query-time cost tractable, but the memory/latency trade-off is still nontrivial \cite{khattab2020colbert}.

\paragraph{Document chunking and granularity.}
RAG systems almost never index full raw documents; instead, they index \emph{chunks}---typically contiguous text windows of some fixed length (e.g., 200--512 tokens) with stride and overlap \cite{chen2017drqa,karpukhin2020dpr}. Chunking controls a precision--recall trade-off:
\begin{itemize}
    \item Smaller chunks: higher precision, because retrieved passages are more focused and can be fed directly to the generator without much irrelevant context.
    \item Larger chunks: higher recall, because a single retrieved unit is more likely to contain the answer somewhere inside it, but it may also contain distracting or noisy text.
\end{itemize}
Overlap (sliding windows with stride $<$ window size) is often used so that information spanning a boundary is not lost.

Chunk granularity interacts with the retriever and the generator. Dense retrievers may benefit from semantically coherent chunks, since embeddings assume each chunk is internally consistent. FiD-style generators \cite{izacard2021fid} can ingest many retrieved chunks independently, so they tolerate smaller, more numerous chunks as long as the index can return enough candidates. In contrast, RAG setups that concatenate top-$k$ passages into a single context window may prefer slightly longer chunks to avoid blowing up the number of retrieved segments.

\subsubsection{Implications for Automated Retrieval Configuration}
The design axes above---sparse vs.\ dense vs.\ hybrid retrieval, late-interaction vs.\ single-vector encodings, reranking strategies, ANN index choice, multi-vector vs.\ single-vector storage, and chunk sizing/overlap---jointly determine recall, precision, latency, memory footprint, and factual grounding quality in a downstream RAG system \cite{karpukhin2020dpr,khattab2020colbert,izacard2021fid}. Critically, these axes are \emph{not} universally optimal: a configuration that works best for noisy, short, user-authored queries over a constantly changing knowledge base may be very different from the configuration that works best for long, technical queries over stable scientific PDFs. Because the retrieval stage gates what evidence the generator ever sees, choosing this configuration is a first-order design problem. This motivates automated frameworks that can search, rank, or adapt retrieval and indexing strategies to the target domain instead of relying on a single hand-tuned recipe.

\subsection{Query Reformulation and Pre-Retrieval Conditioning}
\label{sec:query_reformulation}

The effectiveness of a Retrieval-Augmented Generation (RAG) pipeline depends not only on how we retrieve and index documents, but also on \emph{what} query is actually sent to the retriever. In practice, systems rarely submit the user’s raw question directly. Instead, they apply a set of pre-retrieval conditioning steps---query rewriting, expansion, instruction interpretation, and decomposition---to improve recall, robustness, and controllability. This section surveys these techniques and argues that ``the query that hits the retriever’’ is itself a tunable design parameter that can vary by domain, task, and corpus.

\subsubsection{Query Rewriting and Query Expansion}
\label{sec:query_expansion}

A classic strategy in information retrieval is \emph{query expansion}: adding new terms to the query to increase the likelihood of matching relevant documents. This dates back to pseudo-relevance feedback (PRF), where the system first retrieves an initial set of candidate documents, assumes (some of) them are relevant, extracts salient terms, and then issues an expanded query \cite{rocchio1971relevance,robertson2009probabilistic}. PRF and related relevance feedback methods improve recall by bridging vocabulary gaps between the original query and target passages.

Neural variants of query expansion have adapted this idea to modern retrievers. One line of work uses generation models to synthesize likely answer-bearing text, then uses that synthetic text to guide retrieval. For example, \textit{doc2query} trains a sequence-to-sequence model to generate plausible queries for each document, effectively augmenting the document index with ``imagined'' questions and improving matching for sparse and dense retrievers alike \cite{nogueira2019doc2query}. More recent approaches invert that direction: instead of expanding documents, they expand the \emph{query}. Hypothetical Document Embeddings (HyDE) prompt a large language model to generate a short ``pseudo answer passage'' for the user question; that passage is then embedded and used as the retrieval query in a dense vector index \cite{gao2023hyde}. HyDE improves dense retrieval in cases where the original user query is short, underspecified, or informal, because the generated hypothetical passage provides richer semantic context \cite{gao2023hyde}.

In conversational and multi-turn question answering, query rewriting aims to recover an explicit, context-independent query from anaphoric or underspecified user turns (e.g., ``What about \emph{his} patents?'' or ``And where did \emph{that} happen?''). Systems generate a fully specified standalone query that encodes the conversation history, then submit that rewritten query to retrieval \cite{aliannejadi2020convai,voskarides2020queryrewriting}. This is critical for conversational search and dialogue-style assistants, where raw user turns are often elliptical and cannot be directly matched in the index.

Large language models (LLMs) now make it practical to do on-the-fly query rewriting, self-expansion, or self-clarification at inference time. For example, black-box retrieval-augmented generation systems such as RePlug treat the retriever as a modular component and use iterative query reformulation to surface better supporting evidence before final answer generation \cite{shi2023replug}. This LLM-driven reformulation can inject synonyms, paraphrases, or domain terms that the original user did not explicitly mention, improving recall@k in downstream RAG.

Overall, query rewriting / expansion increases the chance that at least one relevant chunk appears in the retrieved set, which directly benefits the generator in RAG by exposing it to more grounded evidence.

\subsubsection{Instruction-to-Query Mapping and Search Intent Normalization}
\label{sec:intent_mapping}

In many practical deployments, users do not issue ``search queries’’; they issue \emph{instructions}. For instance:
\begin{quote}
    ``Summarize safety violations in last week's turbine log.'' or
\end{quote}

\begin{quote}
    ``Give me the name who made this claim.''
\end{quote}
These instructions are often imperative, stylistic, or under-specified with respect to retrievable content. The retrieval system, however, expects something closer to a factual entity- and evidence-oriented query.

Instruction-to-query mapping refers to transforming these high-level intents into retrieval-friendly search strings that capture the actual information need. In enterprise QA and domain-specific RAG, this often means:
(1) extracting named entities, time ranges, product identifiers, regulation codes, etc.;
(2) reframing subjective or task-oriented language into an objective evidence request (``list safety violations in turbine log from [DATE RANGE]'');
(3) normalizing noisy or informal user language (e.g., slang, abbreviations, hashtags) into canonical terms that match how knowledge is stored in the corpus.

Work on conversational search and multi-turn QA formalizes this as ``query resolution'': converting a user’s contextual question into an explicit, well-formed retrieval query that captures entities, temporal scope, and disambiguated referents \cite{aliannejadi2020convai}. In open-domain QA, similar techniques have been applied to standardize natural-language questions into retrieval-oriented forms that better align with Wikipedia or web-style evidence \cite{voskarides2020queryrewriting}.

For RAG over noisy domains such as social media, chats, or incident logs, this step is crucial. The underlying corpus may encode events in terse, domain-specific shorthand. A generic LLM-style user question (``what happened with engine 3 yesterday?'') is usually not lexically aligned with those internal notes (e.g., ``ENG3 vib spike at 14:37; auto-shutdown interlock tripped''). Systems that explicitly map user questions to retrieval-style intent---by inserting canonical terms like ``vibration spike,'' timestamps, component names, and system IDs---can dramatically improve first-stage recall in both sparse and dense retrieval. This is, in practice, a tunable knob that must often be tailored per domain.

\subsubsection{Decomposition and Multi-Hop Retrieval}
\label{sec:multihop}

Many queries, especially in knowledge-intensive QA, are inherently multi-hop: answering them requires chaining together multiple pieces of evidence across different passages or entities \cite{yang2018hotpotqa}. A single monolithic query may under-specify these intermediate steps. For example:
\begin{quote}
    ``Was the company founded by the same person who led the Mars mission contract?''
\end{quote}
implicitly asks: (i) who founded the company, (ii) who led the Mars mission contract, and (iii) are they the same person.

Question decomposition addresses this by breaking complex questions into simpler sub-queries that can each be retrieved more easily \cite{min2019multihop,talmor2018breakitdown}. A common pattern is iterative retrieval: retrieve for sub-question 1, extract key entities from the result, then use those entities to form sub-question 2, and so on, accumulating evidence across hops \cite{min2019multihop,yang2018hotpotqa}. 

More recent agent-style methods push this further by letting an LLM reason step-by-step and issue retrieval calls at each step. In ReAct, the model interleaves chain-of-thought reasoning with tool calls (including retrieval), generating intermediate hypotheses and follow-up search queries \cite{yao2022react}. This effectively learns to \emph{decompose} at inference time: the system does not just expand the original query, it plans a sequence of targeted retrieval queries that progressively narrow in on the final answer.

Decomposition benefits RAG because it mitigates a known limitation of single-shot retrieval: if the first retrieval call fails to surface the key bridge entity (e.g., the founder's name), the generator never has a chance to see it. By contrast, iterative multi-hop retrieval explicitly searches for those bridge facts and feeds them forward. This is especially important in domains where information is fragmented across many short notes or where evidence is relational (``X acquired Y, Y hired Z, Z led project Q'').

\subsubsection{Implications for Automatic Query Conditioning}
The techniques above---query expansion and rewriting, instruction-to-intent mapping, and multi-hop decomposition---are all \emph{pre-retrieval knobs}. They change not the retriever itself, but the input the retriever sees. Crucially, these knobs are domain-sensitive:
\begin{itemize}
    \item Conversational assistants benefit from context resolution and anaphora rewriting \cite{aliannejadi2020convai}.
    \item Enterprise incident search benefits from canonicalizing slang and surfacing entity IDs.
    \item Open-domain multi-hop QA benefits from explicit decomposition and iterative retrieval \cite{yang2018hotpotqa,min2019multihop,yao2022react}.
    \item Dense semantic search over short, noisy queries benefits from LLM-generated hypothetical expansions such as HyDE \cite{gao2023hyde}.
\end{itemize}

Because RAG answer quality is upper-bounded by what is retrieved, and retrieval is heavily influenced by how the query is phrased, these conditioning steps are not optional engineering details but central design decisions. Our broader framework treats these query-conditioning choices as part of the configurable retrieval stack: rather than assuming a single ``raw user question $\rightarrow$ retriever'' pathway, we explicitly acknowledge that different domains may require different pre-retrieval strategies to maximize recall and downstream factual grounding.

\subsection{Post-Retrieval Aggregation and Answer Generation}
\label{sec:post_retrieval}

Retrieval-augmented generation (RAG) is often described as a two-stage pipeline:
retrieve relevant evidence, then generate an answer conditioned on that evidence
\cite{lewis2020rag,izacard2021fid}.
In practice, however, the ``generation'' half of the pipeline is itself a rich
design space. How retrieved evidence is filtered, organized, presented to the
model, and enforced at decoding time strongly affects factuality, style,
hallucination rate, and cost.
This section surveys the main classes of post-retrieval control:
(i)~context integration strategies,
(ii)~hallucination mitigation and grounding, and
(iii)~the growing alternative of using long-context LLMs instead of retrieval,
and how that trend interacts with RAG.

\subsubsection{Context Integration Strategies}
\label{subsec:context_integration}

\paragraph{Naive concatenation / prompt stuffing.}
The most direct RAG formulation (sometimes called ``vanilla RAG'') simply
concatenates the top-$k$ retrieved passages with the user query and feeds that
augmented prompt to a generator such as a seq2seq model or an instruction-tuned
LLM \cite{lewis2020rag}.
This strategy treats retrieval as a context expander: the generator's decoder
attends jointly over all retrieved text plus the question.

While simple, this approach couples answer quality to the model's context
budget. As $k$ (and thus total tokens) grows, the model may face attention
dilution: relevant evidence is present somewhere in the prompt, but it competes
with distractors.
This motivates more deliberate curation of which passages
actually reach the generator.

\paragraph{Rerank-then-truncate (evidence curation).}
A widely used refinement is to \emph{over-retrieve} (e.g., dozens of passages),
then apply a stronger reranker---often a cross-encoder that jointly encodes
(query, passage) with a large Transformer such as BERT
\cite{nogueira2019passagererank}---to reorder candidates by estimated
task-specific relevance.
The generator then only sees the top $L$ passages under this reranker, where
$L$ is chosen to fit the model's context window.
Many systems additionally apply redundancy-aware selection such as Maximal
Marginal Relevance (MMR), which trades off relevance and novelty, so that the
final context covers diverse supporting evidence instead of repeating the same
snippet phrased slightly differently \cite{carbonell1998mmr}.
Operationally, this ``rerank-then-truncate'' stage is already a nontrivial
hyperparameter surface: Which reranker? How many passages survive? Do we
optimize for precision (few, high-quality passages) or for recall (broad
coverage for safety-critical QA)?

\paragraph{Fusion-in-Decoder (FiD) and multi-passage conditioning.}
Instead of concatenating passages into a single long encoder input,
Fusion-in-Decoder (FiD) encodes each retrieved passage independently, then fuses
all encoded evidence in the decoder cross-attention \cite{izacard2021fid}.
This architecture lets the generator attend to many pieces of evidence without
forcing them to collide in a single flat token stream, which empirically boosts
open-domain QA performance and robustness to noisy retrieval.
In practice, FiD-like decoders are frequently used as strong RAG baselines
because they relax the tension between recall (retrieve many passages) and
tractable decoding.

\paragraph{Reasoning-aware integration (retrieval $\leftrightarrow$ chain-of-thought).}
Classical RAG assumes a single retrieval step before answer generation.
However, complex or multi-hop questions often require \emph{iterative}
evidence gathering.
Recent work proposes interleaving retrieval with step-by-step reasoning,
so that intermediate hypotheses guide what to fetch next, and the newly
retrieved evidence, in turn, updates the reasoning chain.
For example, ReAct prompts an LLM to alternate between reasoning steps
and tool calls (such as web or corpus search), explicitly writing out what it
knows, what it needs next, and how it will use the retrieved snippet in the
final answer \cite{yao2022react}.
IRCoT (Interleaving Retrieval with Chain-of-Thought) follows a similar
philosophy for multi-hop QA: the model emits partial chain-of-thought (CoT)
rationales, retrieves again conditioned on those partial rationales, and then
continues reasoning with the new evidence \cite{trivedi2022ircot}.
This style of retrieval-aware reasoning reduces the classic failure mode where
a one-shot retriever misses a crucial intermediate fact and the generator
hallucinates a bridge.

\subsubsection{Hallucination Mitigation and Grounding}
\label{subsec:hallucination_grounding}

Even with high-quality retrieval, large language models may still generate
assertions that are unsupported (or even contradicted) by the retrieved
evidence.
A major thread in post-retrieval aggregation research is therefore:
\emph{make the model prove its claims.}

\paragraph{Citation-enforcing generation and attribution.}
One line of work directly couples generation with explicit citations.
Early large ``open-book'' QA systems such as GopherCite trained models to
produce answers \emph{and} quote supporting spans, and even to abstain when
evidence is insufficient \cite{menick2022gophercite}.
Subsequent attribution frameworks argue that model outputs should be
\emph{Attributable to Identified Sources} (AIS): every factual statement in the
answer should be traceable to some cited snippet, and that snippet should
actually support the claim \cite{rashkin2023ais}.
More recent ``locally attributable'' generation goes further by restructuring
decoding into stages: first select fine-grained supporting spans from the
retrieved corpus, then plan and realize each output sentence so that \emph{every
sentence} is paired with concise, sentence-level evidence pointers
\cite{slobodkin2024attribute}.
This moves beyond coarse ``URL-style'' citations toward token-level grounding.

\paragraph{Constrained or guided decoding.}
Another approach is to algorithmically constrain decoding to stay anchored
to retrieved evidence.
Reranking and constrained decoding have been explored in multi-step RAG agents
that repeatedly retrieve, verify, and revise candidate answers, editing or
rejecting unsupported claims instead of emitting them verbatim
\cite{gao2024riches,dhuliawala2024cove}.
These systems increasingly frame answer generation as an iterative
``research-then-edit'' loop: draft an answer, verify each claim against
retrieved sources (possibly with explicit follow-up retrieval),
and patch or delete hallucinated spans before returning the final answer.
Self-RAG operationalizes a related idea end-to-end: a single model learns
when to retrieve, how to incorporate that evidence, and how to critique
(and, if needed, correct) its own draft using special reflection tokens,
yielding measurable gains in factuality and citation quality in open-domain
QA and long-form generation \cite{asai2023selfrag}.

Taken together, these methods treat grounding not as an afterthought but as a
contract: the answer should be auditable, sentence by sentence, against known
sources.
This is especially important for high-stakes domains (biomedicine,
law, finance), where unverifiable claims are unacceptable.

\subsubsection{Long-Context LLMs vs.\ Retrieval-Augmented Generation}
\label{subsec:long_context_vs_rag}

An increasingly popular alternative to retrieval is to skip retrieval entirely
and simply feed the model very large chunks of source text.
Commercial frontier models have rapidly expanded context windows
(e.g., hundreds of thousands of tokens up to the order of a million tokens),
making it technically feasible to ``paste the corpus into the prompt'' and ask
the model to answer directly \cite{anthropic2024claude35sonnet,anthropic2025sonnet4}.
This long-context (LC) paradigm challenges the premise that we \emph{must}
retrieve, rerank, and truncate before generation.

However, two practical issues keep RAG competitive.

First, cost and efficiency: attention over hundreds of thousands of tokens is
computationally expensive, and most API pricing scales (roughly) with tokens.
RAG can aggressively narrow the context to only the most relevant passages,
often yielding similar answers with far fewer tokens processed.

Second, precision and controllability: retrieval acts as a targeted filter,
regularizing the model to condition on high-salience evidence instead of
absorbing an entire noisy corpus.
Empirical head-to-head studies comparing LC models to RAG-style systems find
that LC can outperform RAG on some long-context benchmarks \emph{if} the model
is allowed to read the full source, but RAG remains substantially more
cost-efficient and can approach LC quality when paired with
dynamic routing strategies that decide, per query, whether retrieval alone
is enough or whether to escalate to a full long-context read
\cite{li2024selfroute}.
This hybrid view---``RAG first, escalate to LC only when needed''---suggests
that post-retrieval generation is not a solved one-size-fits-all problem.
Rather, it is a decision policy conditioned on domain, latency budget,
and acceptable risk of hallucination.

The generator side of RAG is highly configurable.
Systems differ in \emph{how} they admit evidence (FiD vs.\ concatenation vs.\
iterative CoT-driven retrieval), \emph{how} they police factuality (citation
enforcement, attribution-first planning, constrained decoding, self-critique),
and even \emph{whether} they rely on retrieval at all versus routing to a
long-context model.
Therefore, ``optimal RAG'' cannot be a single static architecture.
Instead, it is an orchestration problem: given a domain and a query, choose
\emph{which} aggregation strategy, factuality control, and context budget to
invoke.

\subsection{Domain Adaptation and Specialized RAG}
\label{sec:domain_adaptation}

A central assumption behind retrieval-augmented generation (RAG) is that one can
``just'' retrieve relevant context and condition a generator on it
\cite{lewis2020rag,izacard2021fid}.
In practice, this assumption fails uniformly across domains:
the retrieval model, the indexing granularity, the chunking strategy, the negative
sampling regime, and even how the final answer is phrased are all tuned
differently for noisy social media logs, corporate knowledge bases, biomedical
literature, or multilingual low-resource collections.
Empirically, RAG is not a single pipeline; it is a family of \emph{domain-shaped} design
choices \cite{khattab2020colbert,izacard2022contriever,zhang2021mrtydi}.
This section surveys how RAG configurations systematically diverge across four
representative settings:
(i) short, informal text,
(ii) formal/encyclopedic material,
(iii) low-resource and non-English languages,
and (iv) proprietary/internal corpora such as enterprise or biomedical QA.
We argue that the community currently discovers effective pipelines
by manual, domain-specific intuition, rather than via a principled, automated search.
This motivates the need for frameworks that \emph{learn} which RAG variant is optimal
for a given domain, especially for languages and sectors that lack expert-curated recipes.

\subsubsection{RAG for Noisy, Informal, Short Text}
\label{subsec:noisy_text}

Many real deployments of RAG operate not over Wikipedia, but over chat transcripts,
customer support tickets, incident reports, Slack channels, call-center notes,
and social media posts.
These sources differ from canonical IR benchmarks along several axes:
utterances are short, ungrammatical, code-switched, and referential
(``it broke again after the last push''); entities are local and time-sensitive
(product SKUs, internal codenames); and discourse is fragmented across turns.
Classical dense retrievers trained on encyclopedic QA
(e.g., DPR) degrade under this distribution shift because the relevance signal
in chat-like corpora is often pragmatic (conversation state, speaker identity)
rather than purely lexical or encyclopedic \cite{karpukhin2020dpr,thakur2021beir}.

Practitioners compensate with domain-tuned adaptations:
(1)~Custom negative sampling that treats adjacent conversation turns as ``hard negatives,''
forcing the retriever to discriminate between two nearly identical utterances differing
only in the specific failure mode or timestamp.
(2)~Windowed or conversational chunking, where the retrieval unit is not a single
message but a sliding block of $k$ turns to preserve local context
(also common in dialogue grounding and multi-turn customer support QA).
(3)~Hybrid sparse+dense retrieval,
where BM25 surfaces exact lexical matches to product/version strings and a dense encoder
handles paraphrastic complaints, then late-fusion ranks them jointly
\cite{thakur2021beir,lin2021pyserini,maillard2021multi}.

Answer generation is also specialized: instead of encyclopedic ``factual'' style,
the target is often task-oriented (``Tell the user to reboot and open Settings''),
and organizations frequently enforce grounded response templates to avoid
hallucinated promises in support channels.
Accordingly, post-retrieval control often includes citation-style or even
verbatim-span quoting from internal support playbooks, plus refusal logic if no
relevant policy snippet is found (to avoid agents inventing policy)
\cite{thoppilan2022lamda,ouyang2022instructgpt}.
In short, RAG in noisy, informal domains emphasizes conversational-window retrieval,
SKU/time-aware lexical hooks, and policy-constrained decoding; this is already
a very different profile than ``vanilla RAG'' on Wikipedia.

\subsubsection{RAG for Formal / Encyclopedic / Regulatory Text}
\label{subsec:formal_text}

When the corpus is long-form, well-edited, and mostly self-contained
(e.g., Wikipedia, manuals, SOP documents, standards, compliance policies,
technical PDFs), the failure modes change.
Documents are longer, internally coherent, and terminology is relatively stable.
Here, the main bottleneck is not noise but \emph{granularity}:
how to segment dense, formally written material into retrievable units
without either (i) diluting relevance with huge chunks or
(ii) losing cross-sentence reasoning needed for correct answers
\cite{izacard2021fid,khattab2020colbert,ram2023incontext}.

A standard tactic in these settings is hierarchical or structured chunking:
split long manuals or policies into semantically coherent sections
(e.g., subsubsection-level or paragraph-level), index those sections,
and then retrieve multiple sections per query. Retrieved sections are then
reranked (e.g., using ColBERT late interaction) and either concatenated or
fused via a generator architecture such as Fusion-in-Decoder (FiD),
which encodes each passage independently and lets the decoder attend across them
\cite{izacard2021fid,khattab2020colbert}.
FiD-like aggregation is particularly effective for dense, reference-style corpora,
where relevant evidence may be scattered over multiple distant paragraphs
(e.g., ``safety override'' is described in \S2.3 and exceptions in Appendix~B).

In compliance and policy QA, another common layer is \emph{grounded justification}:
the answer must quote or cite the specific clause of the regulation for auditability.
As a result, decoding is often constrained or guided to copy normative language
from retrieved passages and to output inline citations \cite{choi2021ethical,
narayan2022plausible}.
These domains therefore bias toward: domain-aware segmentation of long PDFs,
passage-level fusion architectures, and legally/auditably grounded decoding.

\subsubsection{RAG in Low-Resource and Non-English Settings}
\label{subsec:low_resource}

Much of the early RAG literature is implicitly Anglocentric: DPR, FiD, and many
popular instruction-tuned LLMs assume English Wikipedia-scale supervision
\cite{lewis2020rag,karpukhin2020dpr,izacard2021fid}.
However, for low-resource or mid-resource languages (e.g., Turkish, Swahili, Bengali),
the situation is qualitatively different:
(i) publicly available corpora are smaller or noisier,
(ii) pretrained dense retrievers may not cover the language at all,
and (iii) morphologically rich languages (including Turkish) exhibit
agglutinative surface forms that hurt naive lexical matching
\cite{zhang2021mrtydi,zhang2022miracl,reimers2020sentencebert}.

Two broad adaptation strategies appear in multilingual / low-resource RAG:

\textbf{Cross-lingual and multilingual retrievers.}
Multilingual bi-encoders such as mDPR and mContriever are trained via
translation pairs, parallel QA, or distillation from English retrievers
into other languages, allowing semantically aligned dense embeddings across
dozens of languages without per-language supervision
\cite{oguz2021domain,izacard2022contriever,asai2021xorqa}.
Benchmarks like Mr.TyDi and MIRACL explicitly evaluate retrieval in typologically
diverse, often low-resource languages \cite{zhang2021mrtydi,zhang2022miracl}.
These works show that (a) training purely on English data and zero-shot transferring
often underperforms, and (b) modest in-language supervision (or even high-quality
machine translation of questions) can substantially boost recall,
indicating that retrieval itself must be language-aware.

\textbf{Chunking and morphological normalization.}
In agglutinative languages such as Turkish, the same semantic unit can appear with
many inflectional variants, so purely sparse retrieval (BM25) may fragment evidence.
A common workaround is to either stem/lemmatize aggressively, or to rely more heavily
on dense retrieval that abstracts away morphology
\cite{reimers2020sentencebert,zhang2021mrtydi}.
At the same time, Turkish and similar languages often exhibit freer word order,
so naive heuristics for ``question \textrightarrow{} answer sentence'' alignment
transfer poorly from English.
This affects both indexing (how you break passages) and answer extraction
(what you highlight or copy back into the response).

Critically, for these settings there is rarely an off-the-shelf, community-agreed
``best practice'' RAG recipe.
Teams resort to ad hoc mixtures:
translate queries to English, retrieve in English, translate back;
or train a lightweight bilingual retriever on a small in-house parallel QA set;
or bias chunking toward shorter spans to compensate for morphology-heavy tokens.
This points to a gap:
multilingual and low-resource RAG still depends on artisanal, language-specific hacks,
rather than a principled search over design choices.

\subsubsection{RAG over Proprietary / Internal Corpora}
\label{subsec:proprietary}

Enterprise search, legal QA, biomedical and clinical QA, and internal engineering
wikis all share two structural properties:
(i)~their corpora are private and often volatile,
(ii)~the required answers are high-stakes (compliance, patient safety, incident response).
This combination makes generic public RAG recipes brittle.

\textbf{Enterprise \& internal documentation.}
Corporate knowledge bases contain heterogeneous artifacts:
Markdown specs, Jira tickets, long PDF reports, architecture diagrams, postmortems,
meeting transcripts.
These artifacts evolve rapidly and are sometimes mutually contradictory.
Thus, retrieval cannot rely purely on static offline indexing; instead,
pipelines often integrate recency-aware filters (most recent incident runbook first),
access control filters (only retrieve docs the current user can see),
and role-specific answer style (e.g., a summary for executives vs. a step-by-step fix
for on-call engineers).
Hybrid sparse+dense retrieval plus heavy reranking is common here
because internal acronyms and code names are poorly covered in general-domain
embeddings \cite{thakur2021beir,lin2021pyserini,maillard2021multi}.

\textbf{Biomedical / scientific QA.}
Biomedical RAG faces extremely long, jargon-dense PDFs (papers, trial reports) and
a need for verbatim grounding:
models are often required to cite PubMed IDs and cannot ``hallucinate'' mechanisms
of action \cite{jin2019pubmedqa,gu2021pubmedbert,logan2021scent}.
Domain-specialized encoders such as BioBERT, PubMedBERT, SciNCL, or BioLinkBERT
are explicitly trained on biomedical abstracts and clinical notes to improve retrieval
recall over PubMed-scale literature \cite{lee2020biobert,gu2021pubmedbert,gao2022scincl,yasunaga2022linkbert}.
Downstream generation is further constrained by factuality filters and safety rails
(e.g., ``do not provide treatment advice''), which are typically wired in at
decoding time rather than left to general-purpose instruction tuning
\cite{jin2019pubmedqa,yasunaga2022linkbert}.

\textbf{Legal / regulatory QA.}
In legal RAG, ``hallucinating'' a precedent or misquoting a clause can be sanctionable.
Accordingly, systems often force extractive answers: the model must produce
a span copied from retrieved statutes or case law, optionally followed by a short
natural-language paraphrase.
Retrievers for this domain are commonly fine-tuned on statute-to-clause relevance
and trained to respect jurisdictional boundaries (federal vs. provincial, etc.),
which is information absent from generic retrievers
\cite{zhong2020jec,chalkidis2021lexglue}.

Across these proprietary domains, we again observe bespoke recipes:
recency-aware indexing, ACL-aware retrieval, domain-specialized encoders,
and legally/auditably constrained decoding.

\subsubsection{Why This Matters for Automated RAG Design}
\label{subsec:why_framework}

Across all four settings, practitioners do \emph{not} simply reuse a ``reference RAG''
implementation from the literature.
Instead, they iteratively hand-tune:
\begin{itemize}
    \item How to use chunks / window documents (sliding conversational windows vs.
    semantically segmented PDF sections vs. clause-level statutes).
    \item How to post-filter and rerank (recency, access control, policy compliance).
    \item How to force grounding at generation time (citation copying in compliance,
    extractive spans in legal and biomedical, refusal when no policy exists).
\end{itemize}
These knobs are \emph{domain-specific} because the failure modes are
domain-specific: informality and coreference in chat logs, long-tail jargon in
biomed, morphology and code-switching in Turkish, auditability in policy QA,
hierarchical structure in manuals.

However---and this is the gap our work addresses---there is
currently no unified, data-driven framework that \emph{automatically discovers}
the best RAG configuration for a new corpus.
State-of-practice is still largely heuristic:
an engineer guesses a chunk size, picks BM25~+~some dense retriever,
bolts on FiD or naive concatenation, and hopes it generalizes.
This is especially limiting in under-resourced languages (e.g., Turkish) and in
siloed proprietary domains, where there is neither extensive prior art
nor abundant supervised labels to guide intuition
\cite{zhang2021mrtydi,zhang2022miracl,gu2021pubmedbert}. In summary, domain adaptation in RAG today is manual, fragmented, and expert-driven.

\subsection{RAG Tooling, Orchestration Frameworks, and Auto-RAG Systems}
\label{sec:tooling}

Retrieval-augmented generation (RAG) has moved from a research pattern to an engineering product primitive. As a consequence, there is a rapidly growing ecosystem of toolkits, orchestration libraries, and ``auto-RAG'' systems that aim to (i) make it easier to assemble RAG pipelines, and/or (ii) partially automate their tuning. This section situates these developments, then positions our framework as a more general, data-driven optimizer of entire RAG pipelines across domains.

We organize prior work into three trends:
(1)~automatic RAG selection and tuning systems that try to search or adapt a RAG configuration to a task;
(2)~agentic, feedback-driven loops in which the LLM itself critiques retrieval and dynamically adjusts how it retrieves.
We then explain how our framework differs from these works.

\subsubsection{Automatic Module Search, Retriever--Generator Pairing, and Prompt / Pipeline Tuning}
\label{subsec:auto_rag_search}

Beyond ``lego block'' assembly, a parallel line of work tries to \emph{automate} parts of RAG design itself. We group these under automatic RAG configuration search and automatic evaluation / feedback for improvement.

\paragraph{AutoRAG-style pipeline search.}
Recent work explicitly frames RAG construction as a search or optimization problem. The AutoRAG framework proposes to ``automatically identify suitable RAG modules for a given dataset,'' motivated by the observation that retrieval and generation components that work for one domain (e.g., Wikipedia QA) may fail in another (e.g., domain-specific documents) \cite{autorag_kim_2024}. AutoRAG treats the pipeline as a set of interchangeable modules (retriever type, ranker, generator prompting, etc.) and explores combinations to find high-performing variants without requiring the practitioner to manually trial every option \cite{autorag_kim_2024}. This directly mirrors our thesis that RAG is not one-size-fits-all and must be specialized per corpus.

There is also work called ``Auto-RAG'' (sometimes hyphenated) that focuses on \emph{iterative retrieval orchestration}: instead of fixing retrieval to a single pre-answer step, an LLM actively plans and refines multiple retrieval calls, deciding what to fetch next, when to stop, and how to integrate the evidence into a final answer \cite{autorag_yu_2024,autorag_repo_2024}. This variant explicitly models retrieval as a loop between the LLM and the retriever, where the model autonomously reformulates queries, evaluates usefulness of retrieved passages, and terminates when it judges that it has enough evidence \cite{autorag_yu_2024,autorag_repo_2024}. While this work automates \emph{query planning and retrieval depth} at inference time, it does not claim to benchmark or rank fundamentally different retrievers, chunking policies, or generators for a new dataset; instead, it teaches the model to drive an existing retrieval stack more intelligently \cite{autorag_yu_2024}.

\paragraph{Programmatic prompt / pipeline optimization (DSPy).}
DSPy is a declarative framework that treats an LLM pipeline---including RAG pipelines---as a modular program with tunable parameters instead of ad hoc prompts \cite{dspy_site_2025,dspy_repo_2025}. DSPy provides ``optimizers'' that automatically adjust prompt templates and, in some cases, lightweight model parameters to maximize a user-defined metric such as answer accuracy or factuality \cite{dspy_optimizers_2025,dspy_stanford_2024}. In other words, you specify \emph{what} you want (a QA module grounded in evidence, a retriever call followed by a summarizer, etc.) plus a metric, and DSPy searches prompt variants and module wiring to improve that metric, rather than relying on manual prompt tinkering \cite{dspy_site_2025,dspy_optimizers_2025}. This line of work automates prompt engineering and sometimes retriever--generator pairing for a task, but typically assumes a fixed overall pipeline structure (e.g., ``retrieve then answer'') and a single task/domain at a time \cite{dspy_site_2025,dspy_stanford_2024}.

\paragraph{Automated evaluators and diagnostic feedback.}
A related ecosystem aims to \emph{evaluate} and \emph{debug} RAG pipelines with minimal human labeling. RAGAS (Retrieval-Augmented Generation Assessment) proposes largely reference-free metrics---faithfulness, context relevance, answer relevance, etc.---computed by LLMs themselves to estimate whether answers are grounded in retrieved evidence \cite{ragas_es_2024,ragas_docs_2025}. Subsequent work has adapted and stress-tested these metrics in applied domains such as telecom QA, highlighting both their promise and their brittleness under domain shift \cite{ragas_telecom_2024}. More recently, RAGXplain argues that evaluation should not stop at scores: it converts evaluation signals into \emph{actionable guidance}, surfacing where the pipeline is failing (retrieval, ranking, synthesis, etc.) and recommending concrete fixes \cite{ragxplain_2025,ragxplain_pdf_2025}. These systems move toward ``closed-loop'' optimization: measure, diagnose, suggest improvements.

\smallskip
\noindent
\emph{Summary.} Automatic RAG tooling already acknowledges that (i) different corpora demand different retrieval/generation behavior, and (ii) LLM-based judges can guide tuning \cite{autorag_kim_2024,ragas_es_2024,ragxplain_2025}. But existing systems usually optimize one slice at a time: iterative retrieval control \cite{autorag_yu_2024}, prompt templates \cite{dspy_site_2025}, or evaluator-driven suggestions \cite{ragxplain_2025}. They rarely provide a \emph{systematic, corpus-driven search across the full pipeline design space} (chunking granularity, hybrid vs.\ dense retrievers, rerankers, generator style, etc.) and then \emph{rank} those pipelines with objective task metrics.

\subsubsection{Agentic and Feedback-Loop Retrieval Control}
\label{subsec:agentic_feedback}

A third thread teaches the model to reason about its own retrieval. Instead of assuming a static retriever that runs once, these methods let the LLM introspect: ``Do I need to retrieve? Did I retrieve the right thing? Do my claims match the evidence?''

\paragraph{Self-RAG / self-reflective RAG.}
Self-RAG (``Self-Reflective Retrieval-Augmented Generation'') explicitly trains an LLM to (i) decide \emph{when} retrieval is necessary, (ii) request additional evidence mid-generation, and (iii) critique its own output for factual support \cite{selfrag_asai_2023,selfrag_site_2025}. The model emits special control tokens (e.g., to trigger retrieval, to judge whether a passage is relevant, to assess whether a claim is supported) and uses those signals both to ground its answer and to provide inline ``is this supported?'' self-checks \cite{selfrag_asai_2023,selfrag_site_2025}. Empirically, Self-RAG improves factuality and citation accuracy over vanilla RAG baselines and even outperforms larger instruction-tuned models that lack retrieval \cite{selfrag_asai_2023,selfrag_site_2025}. Conceptually, Self-RAG blurs the boundary between ``retriever'' and ``generator'': retrieval is no longer a pre-processing step, but a tool the model can call and then audit.

\paragraph{Agentic RAG in orchestration libraries.}
Modern orchestration frameworks have begun to expose similar behavior as ``agentic RAG'': an agent (LLM) has retrieval as a callable tool and can iteratively reformulate queries, gather more evidence, and only then answer \cite{langchain_agentic_rag_2025,autorag_yu_2024}. This is structurally close to Self-RAG, but usually implemented without specialized fine-tuning: the agent is instructed (via prompting) to reason step-by-step, retrieve when uncertain, and verify answers against retrieved context \cite{langchain_agentic_rag_2025}. The key point is the feedback loop: retrieval quality is not treated as fixed, but is adaptively critiqued and expanded by the model itself \cite{langchain_agentic_rag_2025,selfrag_asai_2023}.

\smallskip
\noindent
\emph{Summary.} Agentic/self-reflective approaches automate \emph{inference-time control}: they let the model itself decide how aggressively to retrieve, how to refine queries, and whether its own claims are grounded \cite{selfrag_asai_2023,langchain_agentic_rag_2025,autorag_yu_2024}. They do \emph{not}, however, automatically choose the \emph{underlying} retriever family, chunking policy, or reranker for each domain; those architectural decisions are still assumed to be fixed and correct.

\subsection{Positioning Our Framework}
\label{subsec:positioning}

Across these strands, we observed the following fundamental limitation. 

Automated/agentic work addresses only pieces of the search problem. AutoRAG-style methods explore module choices or let the model iteratively steer retrieval \cite{autorag_kim_2024,autorag_yu_2024}. DSPy-style optimizers tune prompts and module wiring to maximize a metric \cite{dspy_site_2025,dspy_optimizers_2025}. Self-RAG and agentic RAG teach models to critique retrieval at inference time and request better evidence \cite{selfrag_asai_2023,langchain_agentic_rag_2025}. RAGAS and RAGXplain move toward automated evaluation and actionable debugging of RAG systems \cite{ragas_es_2024,ragxplain_2025}.

What is \emph{still} missing is a unified, dataset-driven \emph{discovery} framework that:
(i) systematically enumerates and instantiates \emph{entire} RAG pipelines---including Pre-Embedding, Query Expansion, Retrieval, Reranking, Passage Filtering, Passage Augmentation, Passage Compression, Prompt Making and Post-Generation;
(ii) runs these candidate pipelines on a target corpus \emph{in that corpus's domain and language};
(iii) scores them with objective metrics (retrieval metrics, LLM-Judge, semantic score); and
(iv) produces an explicit ranking of which pipeline works best for that corpus.

Our framework is built exactly for that role. Unlike generic orchestration libraries, it treats RAG pipeline design itself as an empirical search problem over a \emph{multi-stage space} (nine different technique families), and then reports which combinations dominate on that dataset. This explicit, data-grounded model selection across different domains is, to our knowledge, not provided by existing libraries or self-tuning approaches \cite{autorag_kim_2024,selfrag_asai_2023,ragxplain_2025}.

Despite substantial progress in retrieval-augmented generation (RAG), modern systems remain fragmented: individual works improve retrieval quality \cite{karpukhin2020dense,khattab2020colbert,izacard2021fid}, stabilize the generator through grounding and citation enforcement \cite{lewis2020rag,shuster2021retrieval,gao2023rarr}, or introduce query rewriting and iterative self-refinement loops that help the model ask better questions of the retriever \cite{mahowald2023active,chen2023react}. Other efforts focus on developer-facing RAG tooling and orchestration frameworks, lowering the barrier to composing pipelines but largely assuming that practitioners will hand-tune chunking, retrieval mode, filtering, and answer synthesis for their own data. Critically, none of these threads provides an end-to-end, domain-aware search procedure that, given a specific corpus, \emph{empirically discovers} the optimal RAG configuration. The framework we propose is designed to fill exactly that gap: it treats RAG not as a static architectural choice, but as a controlled design space to be searched per domain.

%% file: sections/03-methods.tex
\section{Method}
\subsection{Problem Statement and Scope}
We formalize the modular RAG design by selecting one technique (or not selecting any technique) from each of the nine RAG technique families to maximize an overall evaluation objective. Let the search space be a product of discrete sets with conditional dependencies across families.

\begin{figure}[h]
    \centering
    \includegraphics[width=1.02\textwidth]{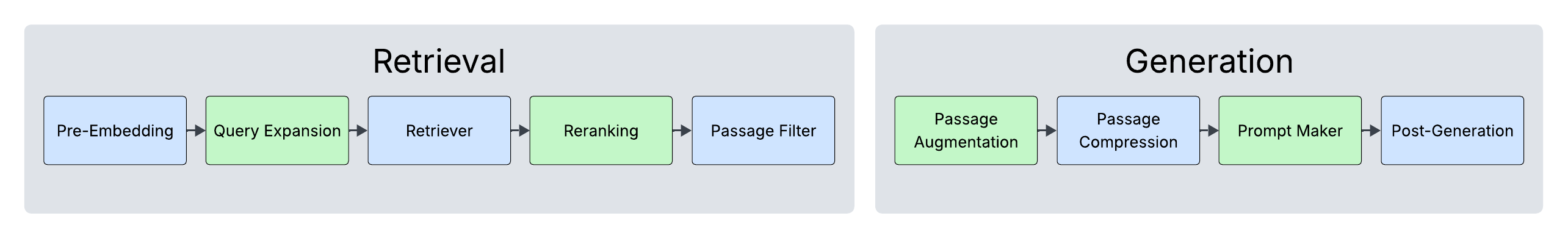}
    \caption{RAG Technique Categories}
    \label{fig:rag_technique_categories}
\end{figure}

\begin{figure}[h]
    \centering
    \includegraphics[width=1.01\textwidth]{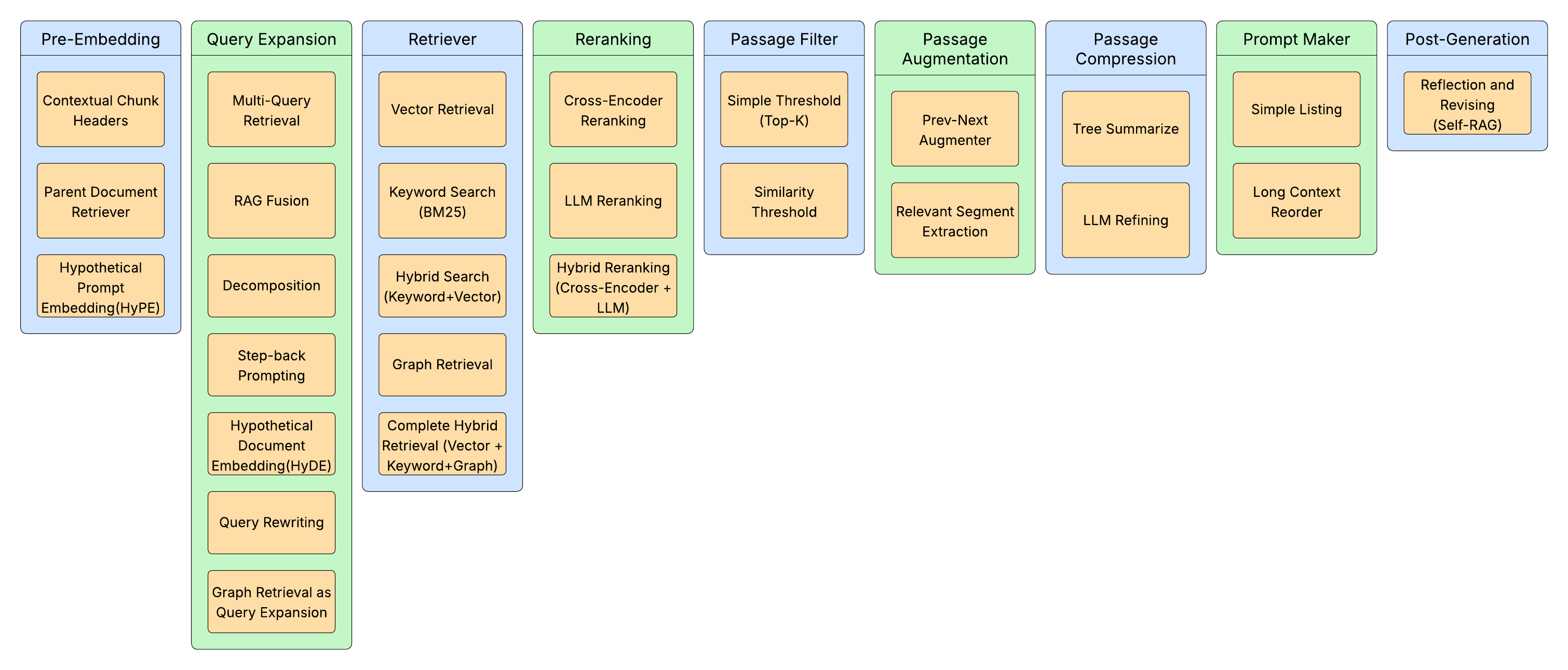}
    \caption{All RAG Techniques used in \method}
    \label{fig:all_rag_techniques}
\end{figure}

\subsection{Modular RAG Design Space (Concrete Families)}
We factor the pipeline into nine technique families implemented in the repository:

\subsubsection{Pre-Embedding}
Pre-Embedding is a process where the chunks of a document will be processed and transformed into a form that is better suited for RAG system, before embedding into a vector database or other retrieval methods.
\begin{enumerate}[leftmargin=*,itemsep=2pt,topsep=0pt]
  \item \textbf{Contextual Chunk Headers}: AutoContext introduces the concept of Contextual Chunk Headers (CCH), a method that enhances the semantic quality of embeddings by attaching document-level and section-level context to each chunk before it is embedded. By prepending these contextual headers, the embeddings capture not only the local meaning of a passage but also its broader role within the document. This enriched representation leads to notable improvements in retrieval quality, increasing the likelihood of retrieving the correct information while reducing the presence of irrelevant results. As a result, CCH improves both the precision of retrieval and the reliability of downstream applications, mitigating instances where large language models might otherwise misinterpret or misapply isolated text fragments.\cite{dsRAG}
  \item \textbf{Parent Document Retriever}: The Parent Document Retriever addresses the trade-off between semantic precision and contextual coherence in document chunking. While smaller chunks yield highly specific embeddings, they often lose connection to the broader narrative of the source text. Conversely, larger chunks preserve context but dilute the semantic relevance of fine-grained details. This method resolves the tension by embedding small, semantically focused chunks while maintaining links to their originating parent documents. During retrieval, the system surfaces both the granular chunks and their associated parent content, thereby providing precise semantic matches supplemented with broader contextual grounding. In doing so, the Parent Document Retriever enhances retrieval quality by balancing fine-grained relevance with document-level coherence. \cite{10962837}
  \item \textbf{Hypothetical Prompt Embedding}: HyPE is a retrieval enhancement method that improves alignment between user queries and relevant content in RAG systems. It operates by precomputing multiple hypothetical prompts for each document chunk during the indexing stage and embedding these prompts into the retrieval space. At inference time, user queries are matched against these stored prompts, effectively reframing retrieval as a question-to-question task. This approach strengthens semantic alignment, reduces retrieval errors caused by style mismatches, and avoids additional computational costs during query execution. Experimental results show that HyPE achieves significant gains in both precision and recall, while remaining compatible with other retrieval strategies, making it a scalable and efficient solution for large-scale RAG pipelines.  \cite{vake2024bridging}
\end{enumerate}

\subsubsection{Query Expansion}
Query Expansion is a process where the user query is transformed or expanded to get a better query or wider query pool to retrieve more related chunks during retrieval.
\begin{enumerate}[leftmargin=*,itemsep=2pt,topsep=0pt]
  \item \textbf{Multi-Query Retrieval}: The multi-query retriever is designed to strengthen information retrieval by generating several distinct queries for a single task. Rather than relying on one perspective, this approach captures content from different angles, reducing the risk of overlooking relevant details. Because the queries are produced through large language models, they remain contextually consistent with the user’s intent, making this method particularly valuable for complex or multi-dimensional questions. \cite{langchain_multqueryretriever}
  \item \textbf{RAG Fusion}: RAG Fusion integrates retrieval-augmented generation with reciprocal rank fusion to improve the ranking and selection of retrieved documents. By issuing multiple queries, it draws on a broader set of candidate results, which are then re-ordered through a rank aggregation process. The use of reciprocal rank values ensures that documents consistently identified as highly relevant across sources receive greater weight, minimizing the bias of any single retrieval method and leading to more reliable outcomes. \cite{rackauckas2024rag}
  \item \textbf{Decomposition}: Decomposition tackles the challenge of large or complex queries by breaking them into smaller, more manageable components. Each sub-question can be addressed with greater precision, and the results are later combined into a coherent overall response. This step-by-step process not only improves the accuracy of retrieval but also supports more effective reasoning, as the system can focus on specific aspects of the problem without being overwhelmed by its full complexity. \cite{chan2024rq}
  \item \textbf{Step-back Prompting}: Step Back Prompting is a prompting strategy designed to strengthen the abstraction capabilities of large language models. Rather than concentrating on narrow details, the technique encourages the model to “step back” and frame the task in terms of broader principles and underlying concepts. By guiding the model toward high-level reasoning, Step Back Prompting supports more structured and coherent problem-solving and helps prevent it from being sidetracked by irrelevant specifics. This shift in perspective enhances conceptual understanding and allows the model to generate responses that are more accurate, logically consistent, and contextually aligned, particularly when handling complex or cognitively demanding queries. \cite{zheng2023take}
  \item \textbf{Hypothetical Document Embedding}: The HyDE method, introduced by Gao et al. (2022), advances retrieval performance by leveraging the generative capacity of large language models. Instead of relying exclusively on direct query–document matching, HyDE prompts the model to generate a hypothetical answer that represents a plausible response to the query. This synthesized answer is then embedded and used to retrieve documents semantically aligned with the generated text. By acting as a semantic bridge, HyDE reduces dependence on surface-level keyword matches and improves the system’s ability to capture nuanced and contextually rich information. As a result, it produces more precise and semantically meaningful retrieval outcomes while illustrating the growing role of language models in embedding-based information access. \cite{gao2023precise}
  \item \textbf{Query Rewriting}: Query Rewriting enhances retrieval-augmented generation by adapting the input query itself rather than modifying the retriever or reader. The approach introduces a rewrite–retrieve–read pipeline, where the original input is first reformulated into a clearer, more retrieval-friendly query. This reduces the mismatch between the phrasing of user inputs and the knowledge needed for retrieval, thereby improving the quality of retrieved contexts. A small trainable language model can be employed as the rewriter, refined through reinforcement learning with feedback from the LLM reader to better align queries with downstream tasks. Empirical results on open-domain and multiple-choice QA demonstrate consistent performance improvements, showing that query rewriting provides an effective and scalable means of strengthening retrieval-augmented systems. \cite{ma2023query}
  \item \textbf{Graph Retrieval as Query Expansion}: Graph Retrieval will be explained in detail later in Retrieval section. This method utilizes Graph Retrieval as a means to expand the query. In other words, firstly, the query is sent to a Graph Retrieval to obtain most relevant relations in the knowledge graph. Then these relations are considered as expanded queries and used similar to Multi-Query Retrieval. This method is one of the novel Advanced RAG Techniques in our comprehensive pipeline.
\end{enumerate}

\subsubsection{Retrieval}
Retrieval is the main part of the Retrieval component in RAG applications. It aims to retrieve the most relevant document chunks according to user query using various methods.  
\begin{enumerate}[leftmargin=*,itemsep=2pt,topsep=0pt]
  \item \textbf{Vector Retrieval}: Vector retrieval methods represent both queries and candidate documents (or chunks) as continuous embeddings in a shared semantic space, enabling similarity-based ranking based on inner products or cosine scores. By encoding semantics rather than relying on exact token overlap, vector retrieval can capture latent relationships and expand the scope of relevance beyond lexical matches. To scale this to large corpora, efficient data structures and approximate nearest neighbor (ANN) algorithms are used to retrieve top candidates from millions or billions of vectors. Recent work has extended this paradigm to hybrid and multi-vector approaches but at its core, vector retrieval remains a powerful means to bridge the gap between natural language queries and richly encoded textual content. \cite{ZhaoEtAl2022_DenseTextRetrievalSurvey}
  \item \textbf{Keyword Search (BM25)}: BM25 is a classical probabilistic retrieval model that ranks documents according to their estimated probability of relevance to a given query. It refines traditional keyword matching by incorporating term frequency, inverse document frequency, and document length normalization into a simple scoring function. This balance allows BM25 to weight informative terms more heavily while preventing overly frequent terms or long documents from dominating the ranking. Despite the rise of neural retrieval methods, BM25 remains a strong and widely used baseline due to its robustness, interpretability, and consistently competitive performance in information retrieval tasks. \cite{INR-019}
  \item \textbf{Hybrid Search (Keyword + Vector)}: Hybrid Search combines lexical search (e.g., BM25) with semantic search (e.g., transformer-based embeddings) to exploit the complementary strengths of both approaches. Lexical methods are efficient and precise for exact term matching but suffer from vocabulary mismatch, while semantic methods capture deeper meaning but can struggle in out-of
  -domain settings. By fusing their scores—through methods such as convex combinations or reciprocal rank fusion—Hybrid Search achieves more robust and effective retrieval, improving both recall and ranking quality across diverse domains. \cite{10.1145/3596512}
  \item \textbf{Graph Retrieval}: Graph Retrieval enriches retrieval-augmented generation by representing documents and their relationships as a graph structure rather than treating them as isolated chunks. In this framework, nodes may correspond to text passages, entities, or concepts, while edges encode semantic or relational links across the corpus. Retrieval then operates not only on direct text similarity but also on traversals of the graph, enabling multi-hop reasoning and the discovery of context that spans multiple documents. This approach improves the system’s ability to handle complex queries requiring synthesis of dispersed information, thereby offering more coherent and contextually grounded retrieval outcomes. \cite{han2024retrieval}
  \item \textbf{Complete Hybrid Retrieval}: Currently three methods are present in \method. Complete Hybrid Retrieval aims to fuse all these methods —through methods such as convex combinations or reciprocal rank fusion—to obtain a more robust retrieval for adapting various queries and datasets.
\end{enumerate}

\subsubsection{Reranking}
Reranking is a process where the retrieved chunks are reranked according to their relevance to the query as an extra step.
\begin{enumerate}[leftmargin=*,itemsep=2pt,topsep=0pt]
  \item \textbf{Cross-Encoder Reranking}: A cross-encoder reranker refines a preliminary set of retrieved candidates by jointly encoding each query–document pair and producing a relevance score via full cross-attention. Because it models interactions between query and passage tokens in a unified representation, it can capture subtle semantic dependencies that independent encodings miss. This richness in feature modeling makes cross-encoders particularly effective for reranking small sets of candidates with high precision. However, their computational cost scales with the number of pairs, which is why they are typically used only in a second-stage reranking step following an efficient first-stage retriever. Recent work also explores “shallow” cross-encoders to balance latency and effectiveness. \cite{PetrovEtAl2024_ShallowCrossEncoders}
  \item \textbf{LLM Reranking}: LLM-based rerankers employ large language models to refine the ordering of retrieved passages by assessing their relevance to a given query. Instead of relying on additional training or fine-tuning, these approaches typically leverage prompt engineering to guide the model in comparative judgment, ranking candidate passages according to contextual fit and semantic alignment. This makes them flexible and adaptable across domains, as they can harness the reasoning and comprehension abilities of LLMs without the cost of retraining. By directly incorporating natural language understanding into the reranking stage, LLM-based rerankers enhance retrieval pipelines with improved precision and context-awareness. \cite{kim2024autorag}
  \item \textbf{Hybrid Reranking (Cross-Encoder + LLM)}: A Hybrid Reranker integrates both cross-encoder and LLM reranking into a unified pipeline to benefit from their complementary strengths. In such a design, an initial cross-encoder stage provides precise token-level scoring for a narrowed candidate set, while a subsequent LLM stage reorders or refines that ranking using broader reasoning, context awareness, or listwise criteria. By cascading these techniques, hybrid rerankers can achieve higher overall accuracy without incurring the full computational cost of applying LLM reasoning to large candidate sets. Empirical studies show this combined strategy reduces ranking errors, especially in out-of-domain scenarios, and balances precision, flexibility, and efficiency in reranking systems. \cite{DejeanClinchantFormal2023_ComparisonCrossEncLLM}
\end{enumerate}

\subsubsection{Passage Filter}
Passage filter is simply the policy on how to limit the number of chunks that will be sent to the generation component.
\begin{enumerate}[leftmargin=*,itemsep=2pt,topsep=0pt]
  \item \textbf{Simple Threshold (Top-K)}: Simple threshold simply filters the top-k relevant documents out of the retrieved documents according to their retrieval (similarity) score. 
  \item \textbf{Similarity Threshold}: Similarity threshold does not limit the number of retrieved documents, instead it filters the documents according to whether their similarity (retrieval) score exceed a set threshold.
\end{enumerate}

\subsubsection{Passage Augmentation}
Passage augmentation enhances retrieval performance by expanding the initial set of retrieved passages with additional, contextually related content. Using metadata such as relationships between neighboring passages, it identifies and incorporates supplementary text segments to provide a broader and more coherent context for generation.\cite{kim2024autorag}
\begin{enumerate}[leftmargin=*,itemsep=2pt,topsep=0pt]
  \item \textbf{Prev-Next Augmenter}:  Prev-Next Passage Augmenter improves retrieval by including neighboring passages alongside the initially retrieved ones. During chunking, each passage is linked to its preceding and succeeding segments, allowing the system to retrieve contextual neighbors that may hold supplementary information. This method enhances coherence and recall by ensuring that relevant surrounding context is not overlooked during retrieval.\cite{kim2024autorag}
  \item \textbf{Relevant Segment Extraction (RSE)}:Relevant Segment Extraction is a query-time technique that merges clusters of related text chunks into larger, semantically coherent segments. This method aims to provide language models with richer contextual input by identifying the portions of text most relevant to the query, rather than relying on fixed-length chunk boundaries. While straightforward factual queries may be resolved within a single chunk, complex questions often require synthesizing information spread across multiple sections. By dynamically constructing these extended segments, RSE ensures that retrieval captures the full context necessary for accurate and well-grounded generation. \cite{dsRAG}
\end{enumerate}

\subsubsection{Passage Compression}
In Passage Compression, the long passages/chunks that are retrieved are summarized or refined to get smaller and more compressed ones. 
\begin{enumerate}[leftmargin=*,itemsep=2pt,topsep=0pt]
  \item \textbf{Tree Summarize}: Tree Summarize provides a hierarchical method for aggregating retrieved information in RAG pipelines. Instead of processing all retrieved passages at once, the method organizes them into smaller groups and generates summaries for each. These intermediate summaries are then progressively merged in a tree-like structure until a final comprehensive synthesis is produced. By compressing information step-by-step, Tree Summarize makes it easier to handle long contexts while ensuring that the final output reflects evidence from across the entire retrieved set. \cite{llamaindex_treesummarize}
  \item \textbf{LLM Refining}: LLM Refine adopts an iterative strategy for response synthesis in retrieval-augmented generation. The process begins with an initial draft answer constructed from a subset of retrieved passages. As additional evidence is introduced, the model revisits its prior output and refines it, modifying the response only when the new material provides stronger or corrective information. This incremental updating mechanism allows the model to build a more accurate and complete answer without discarding previously valid reasoning. \cite{llamaindex_refine}
\end{enumerate}

\subsubsection{Prompt Maker}
Prompt Maker combines the processed passages/chunks to finally form a context string (or prompt) ready to send to the generator for answer generation.
\begin{enumerate}[leftmargin=*,itemsep=2pt,topsep=0pt]
  \item \textbf{Simple Listing}: Simple listing only concatenates the retrieved and processed chunks to form a list of documents.
  \item \textbf{Long Context Reorder}: Long Context Reorder is designed to counteract the tendency of large language models to emphasize the beginning and end of long inputs while overlooking material in the middle. The method works by repositioning the most relevant one or two passages so that they appear at both the start and the end of the prompt. By reinforcing key content in these high-attention regions, the approach ensures that crucial information is not neglected, leading to more accurate and contextually grounded model outputs. \cite{kim2024autorag}
\end{enumerate}

\subsubsection{Post-Generation}
Post-Generation acts as the final step, post-processing the answer before presenting it to the user.
\begin{enumerate}[leftmargin=*,itemsep=2pt,topsep=0pt]
  \item \textbf{Reflection and Revising (Self-RAG)}: Self-RAG introduces a retrieval framework that incorporates self-reflection into the generation process, allowing models to critique their own outputs. The system signals the need for external evidence, assesses the relevance and support of retrieved passages, and evaluates the overall usefulness of its responses. This mechanism enables the model not only to ground its outputs in reliable sources but also to revise them for greater factual accuracy and coherence. As a result, Self-RAG improves both the precision of retrieval and the trustworthiness of long-form generations across diverse tasks. \cite{asai2024selfrag}
\end{enumerate}

\subsection{Search Space Parameterization}
Families and options are drawn from concrete implementations registered in the repository. Each gene indexes a valid option per family; feasibility checks rule out incompatible combinations (e.g., reranking requiring retrieval outputs of sufficient size). Each gene encodes a categorical decision within the modular RAG pipeline. This representation allows the search to explore heterogeneous architectures spanning multiple retrieval paradigms.

\subsection{Objective and Scalarization}

Candidates are scored by an overall function \(F(\mathbf{x}) \in [0,1]\) combining retrieval metrics (Recall@k, mAP, nDCG@k, MRR) and generation metrics (LLM-based score, semantic similarity). We also report component metrics alongside \(F(\mathbf{x})\).

\begin{align}
\text{RetrievalScore}(\mathbf{x}) &= \frac{1}{4} \left( \text{Recall@k}(\mathbf{x}) + \text{mAP}(\mathbf{x}) + \text{nDCG@k}(\mathbf{x}) + \text{MRR}(\mathbf{x}) \right) \\[6pt]
\text{GenerationScore}(\mathbf{x}) &= \frac{1}{2} \left( \text{LLM-Judge}(\mathbf{x}) + \text{Semantic}(\mathbf{x}) \right) \\[10pt]
F(\mathbf{x}) &= \frac{1}{2} \left( \text{RetrievalScore}(\mathbf{x}) + \text{GenerationScore}(\mathbf{x}) \right)
\end{align}

\begin{algorithm}[htbp]
\caption{\method: Genetic Architecture Search for Modular RAG}
\label{alg:rag_gas}
\begin{algorithmic}[1]
\Require category sizes $(d_1,\dots,d_{9})$, population $P$, generations $T$, rates $(p_c, p_m)$, elitism size $k$
\State Initialize population $S_0$ with random genes $x_i \in \{0,\dots,d_i-1\}$
\State Evaluate $S_0$ via API with caching to obtain $F(\bm{x})$
\For{$t=1$ to $T$}
  \State $E_t \gets$ EliteSelect$(S_{t-1}, k)$ 
  \State $O_t \gets$ UniformCrossover$+$AdaptiveMutation$(E_t, p_c, p_m)$
  \State Evaluate $O_t$ via API with caching
  \State $S_t \gets$ ElitistReplacement$(S_{t-1}, O_t)$
  \If{TargetReached or NoImprovement$\ge$Patience} \textbf{break}
  \EndIf
\EndFor
\State \Return Best individual and run statistics
\end{algorithmic}
\end{algorithm}

%% file: sections/04-experiment.tex
\section{Experiment}
\subsection{General Setup and Purpose}
We present \method, a framework designed to systematically explore combinations of advanced retrieval-augmented generation (RAG) techniques on a given dataset in order to identify the configuration that achieves the highest task performance for that dataset. To search for an optimal configuration, \method~ employs a genetic search algorithm, as described in Section~3.2. In this study, our goals are threefold: (i) to evaluate the effectiveness of \method~ in discovering high-performing combinations of RAG techniques, (ii) to analyze whether the performance gains associated with specific technique combinations exhibit consistent patterns across datasets or instead vary in a subject-specific manner, and (iii) to discover the relation between the question types and RAG configurations.

\subsection{Dataset Characteristics and Question Type Distribution}

Our experimental evaluation was conducted across six domain-specific datasets, each exhibiting distinct characteristics that influenced the performance of different RAG configurations. Table~\ref{tab:dataset_stats} presents comprehensive statistics for each dataset.

\begin{table}[!t]
\centering
\caption{Dataset Characteristics and Statistics}
\label{tab:dataset_stats}
\small
\begin{tabular}{lcccccc}
\toprule
\textbf{Dataset} & \textbf{Chunks} & \textbf{Avg. Chunks/Article} & \textbf{Questions} & \textbf{Tokens} & \textbf{Avg. Tokens/Chunk} \\
\midrule
Mathematics & 401 & 40.1 & 100 & 75,540 & 188.4 \\
Law & 373 & 37.3 & 100 & 67,960 & 182.2 \\
Finance & 435 & 43.5 & 100 & 81,212 & 186.7 \\
Medicine & 560 & 50.9 & 100 & 106,552 & 190.3 \\
Defense Industry & 506 & 50.6 & 100 & 94,189 & 186.1 \\
Computer Science & 344 & 34.4 & 100 & 65,157 & 189.4 \\
\midrule
\textbf{Total} & \textbf{2,619} & \textbf{42.9} & \textbf{600} & \textbf{490,610} & \textbf{187.3} \\
\bottomrule
\end{tabular}
\end{table}

The datasets vary significantly in content density and granularity, reflecting the different nature of Wikipedia articles across domains. Medicine demonstrates the highest content volume (106,552 tokens) and chunk density (50.9 chunks/article), characteristic of comprehensive medical encyclopedia entries covering symptoms, treatments, mechanisms, and related conditions. Computer Science has the most compact representation (34.4 chunks/article, 65,157 tokens), reflecting the more focused, definition-oriented nature of technical computing articles. This variation in dataset characteristics proved instrumental in understanding the efficacy of different RAG technique combinations.

Analysis of question categories across datasets reveals important patterns that correlate with RAG performance improvements. Table~\ref{tab:question_distribution} presents the distribution of question types.

\begin{table}[!t]
\centering
\caption{Question Type Distribution Across Datasets}
\label{tab:question_distribution}
\small
\begin{tabular}{lccccccc}
\toprule
\textbf{Question Type} & \textbf{Math} & \textbf{Law} & \textbf{Finance} & \textbf{Medicine} & \textbf{Defense} & \textbf{Computer Science} & \textbf{Total} \\
\midrule
\textit{factual} & 25\% & 23\% & 19\% & 23\% & 24\% & 30\% & 24\% \\
\textit{interpretation} & 34\% & 49\% & 51\% & 53\% & 46\% & 37\% & 45\% \\
\textit{long-answer} & 41\% & 28\% & 30\% & 24\% & 30\% & 33\% & 31\% \\
\bottomrule
\end{tabular}
\end{table}

Mathematics exhibits the highest proportion of \textit{long-answer} questions (41\%), followed by Computer Science (33\%). Conversely, Medicine (53\%), Finance (51\%), and Law (49\%) demonstrate a predominance of \textit{interpretation} questions. As we demonstrate in subsequent sections, this distribution has significant implications for the effectiveness of different RAG strategies.

\subsection{Baselines}
The baseline condition for the experiment is defined as a naive RAG (i.e., vanilla RAG) configuration instantiated within \method. In this setting, none of the advanced RAG techniques are enabled; the system relies solely on vector-based retrieval and straightforward answer generation. The performance obtained under this configuration for each of the six datasets constitutes the baseline against which all other configurations are compared.

\subsection{Metrics and Evaluation}
For retrieval evaluation, we computed an equally weighted aggregate of mean Average Precision (mAP), normalized Discounted Cumulative Gain at rank (k) (nDCG@k), Recall@k, and Mean Reciprocal Rank (MRR).

For generation evaluation, we used an equally weighted combination of two metrics: (i) a semantic similarity score derived from an embedding model, and (ii) an LLM-based judgment produced by a sufficiently large language model. These two metrics were likewise assigned equal weights.

Beyond separately evaluating retrieval and generation performance, we introduced a unified scalar metric to improve interpretability and to act as the fitness function for the genetic search algorithm. Specifically, the overall score is defined as the mean of the retrieval and generation scores, reflecting their equal importance.

\subsection{Implementation Details}

For the LLM generation and other techniques that require LLM capabilities, we used the latest version of Qwen235b-a22b-Instruct-2507 model. For LLM-Judge model for evaluation, we used gpt-oss-120b model. For embedding operations and similarity-score calculation we used mxbai-embed-large model. For cross-encoder reranking we used BAAI/bge-reranker-v2-m3. For reranking of the passages in dataset creation, we used Cohere API and rerank-v3.5 model. 

\subsection{Genetic Search Run and Implementation Details}

For our framework, a comprehensive genetic search framework was developed to explore combinations of optimization and model components efficiently. The current experimental pipeline contains a total of 46,080 possible combinations of techniques. Exhaustively testing all of these configurations would be computationally expensive; therefore, a genetic algorithm (GA) was employed to guide the search toward promising configurations. On average, each run of the genetic search evaluated approximately 100 unique combinations.

To ensure efficient execution, a caching mechanism was integrated into the search process. Each evaluated configuration was represented as an array encoding the selected options for each component, and stored alongside its corresponding fitness value. For instance, a configuration such as \([1, 0, 5, 2, 0, 1, 0, 2, 0, 1]\) corresponds to a fitness score of 0.8254. The array elements denote the selected indices for each configurable component, while the associated score indicates the resulting fitness value of that configuration. This caching mechanism prevented redundant evaluations of previously tested configurations and significantly improved runtime performance.

The fitness score values shown in the results were computed using the formulations provided in Section 3.4, specifically Equations (1), (2), and (3). The term \( F(x) \) in those equations directly represents the fitness score employed by the genetic search algorithm.

The framework was designed to support a wide range of implementations for each core component of the genetic algorithm, enabling extensive experimentation and flexibility. Specifically:

\begin{itemize}
    \item \textbf{Mutation Operators:} Adaptive Mutation, Random Mutation, Categorical Mutation, Swap Mutation, Inversion Mutation, and Composite Mutation were implemented. These operators vary in how they introduce diversity into the population, ranging from adaptive probability adjustments to categorical and position-based perturbations.
    \item \textbf{Crossover Operators:} The framework supports Single-Point Crossover, Multi-Point Crossover, Uniform Crossover, Order Crossover, and Segment Crossover. These operators differ in how they recombine parent chromosomes to form new offspring, allowing both localized and global genetic mixing to be explored.
    \item \textbf{Selection Methods:} Several strategies were implemented, including Tournament Selection, Roulette Wheel Selection, Rank Selection, and Elite Selection. This diversity allows balancing between exploration and exploitation depending on the desired convergence characteristics of the run.
\end{itemize}

Although the framework supports full customization, the configuration that yielded the most stable and high-performing results across experiments is summarized below. The parameters were chosen after several pilot runs and empirical tuning.

\begin{center}
\begin{tabular}{ll}
\toprule
\textbf{Parameter} & \textbf{Value / Method} \\
\midrule
Population size & 16 \\
Number of generations & 20 \\
Crossover rate & 0.6 \\
Mutation rate & 0.08 \\
Elitism count & 5 \\
Selection method & Elite Selection \\
Crossover method & Uniform Crossover \\
Mutation method & Adaptive Mutation (\( 0.01 \leq \text{rate} \leq 0.2 \)) \\
Convergence threshold & 100 \\
Target fitness & 1.0 \\
Random seed & 42 \\
\bottomrule
\end{tabular}
\end{center}

In this configuration, a population size of 16 and 20 generations were selected to maintain a balance between exploration capability and computational feasibility. The elitism count of 5 ensures that the top-performing solutions are preserved across generations, improving convergence stability. The Uniform Crossover with a probability of 0.6 provided strong performance in maintaining population diversity while retaining useful genetic information from parent solutions. An Adaptive Mutation strategy dynamically adjusted the mutation rate between 0.01 and 0.2 according to population diversity, balancing exploration and exploitation throughout the search. The Elite Selection method consistently led to faster and more stable convergence compared to Tournament or Rank Selection in this setup. Each genetic search run proceeded through 20 generations with an initial population of 16 individuals, resulting in approximately 100 evaluated unique configurations per run. Given the high-dimensionality of the search space (46,080 total combinations), this approach provided a computationally tractable yet highly effective strategy for identifying top-performing configurations. The integrated caching mechanism and modular GA design enabled rapid experimentation with different mutation, crossover, and selection operators, ultimately resulting in a robust and efficient optimization pipeline.

%% file: sections/05-results_and_discussion.tex
\section{Results and Discussion}

\subsection{Main Performance Comparison}

\begin{figure}[!t]
  \centering
  \begin{subfigure}{0.48\textwidth}
    \centering
    \includegraphics[width=\linewidth]{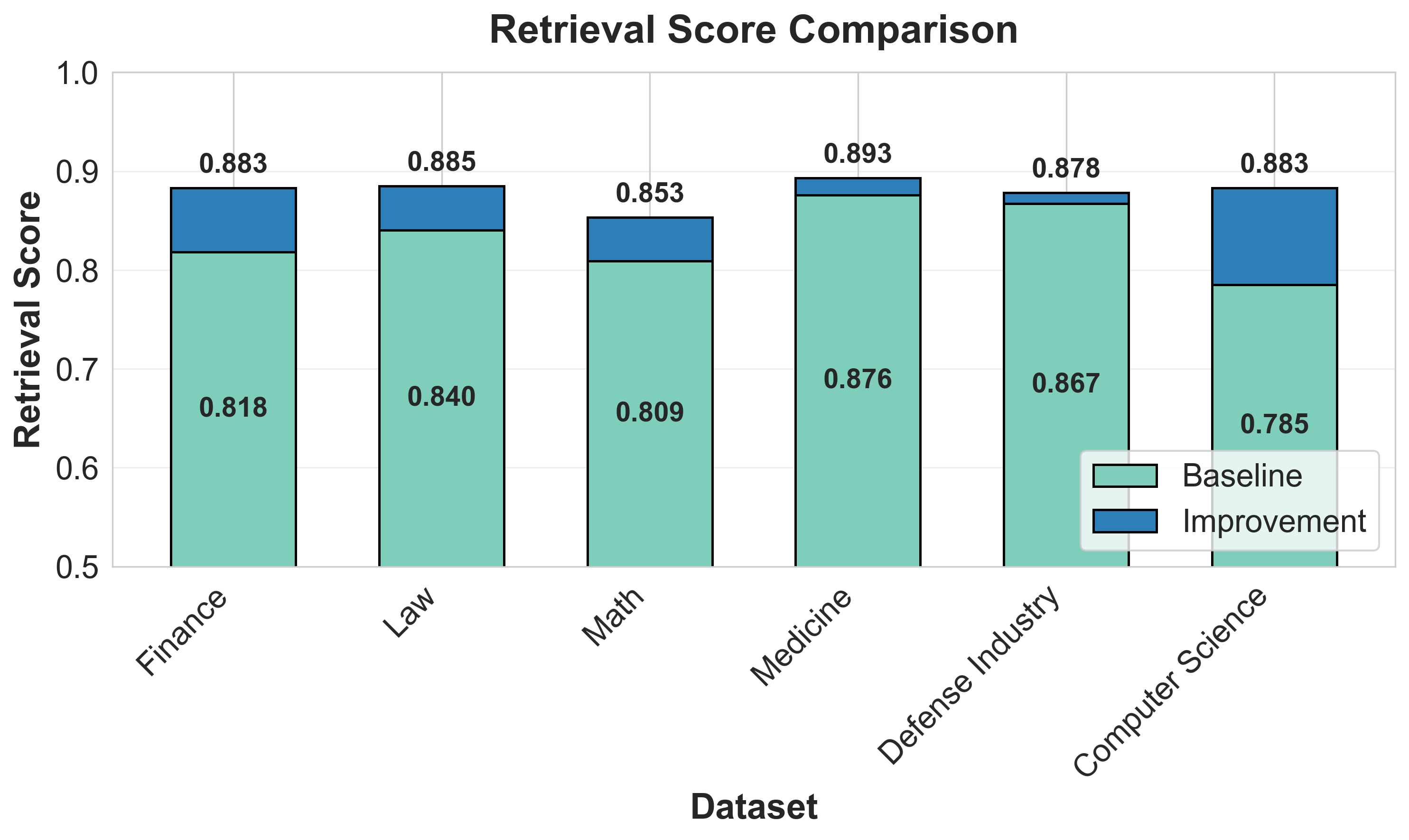}
    \caption{Retrieval performance comparison.}
    \label{fig:retrieval}
  \end{subfigure}\hfill
  \begin{subfigure}{0.48\textwidth}
    \centering
    \includegraphics[width=\linewidth]{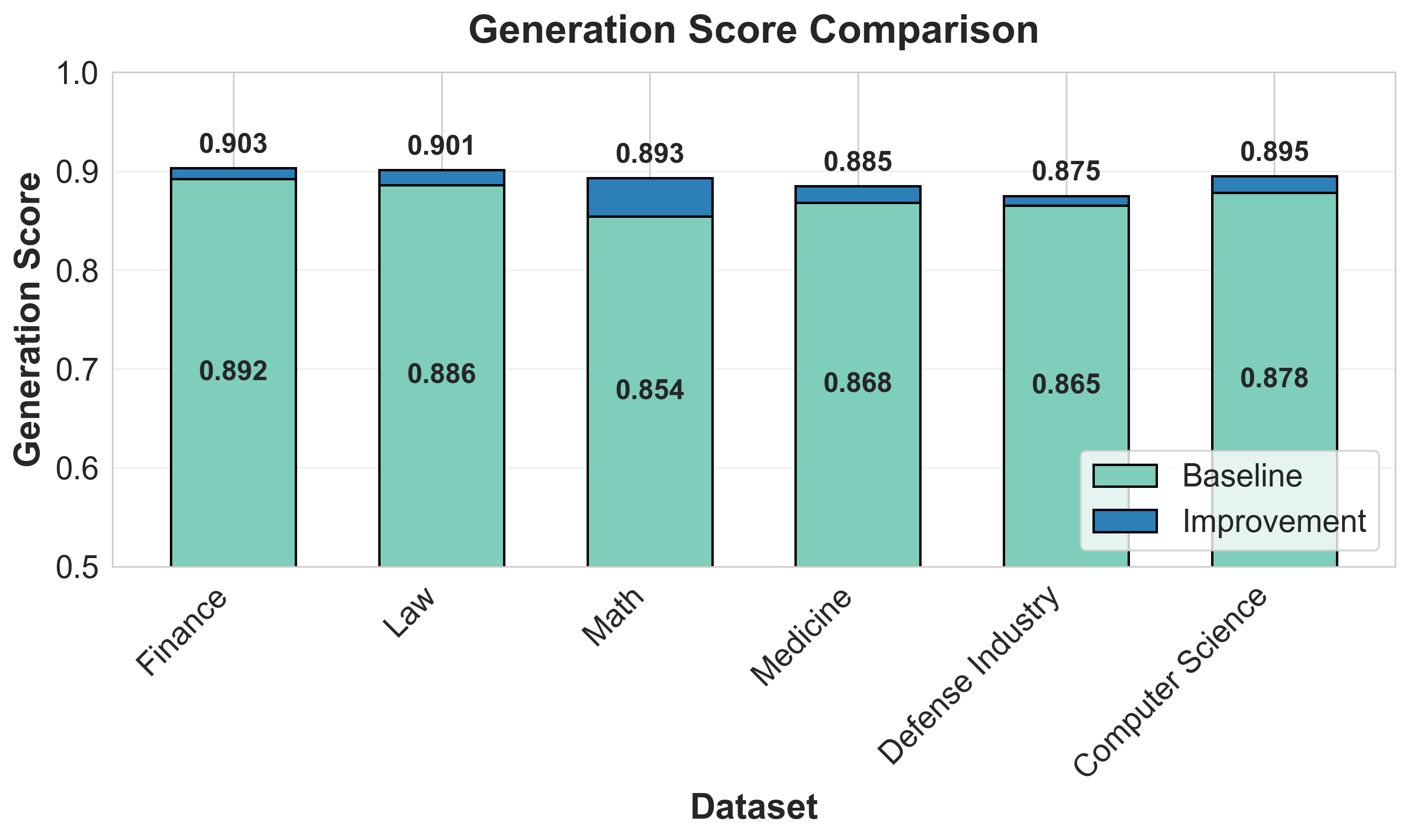}
    \caption{Generation performance comparison.}
    \label{fig:generation}
  \end{subfigure}
  \caption{Retrieval and generation score comparisons across datasets.}
  \label{fig:retrieval_generation}
\end{figure}

\begin{figure}[!t]
  \centering
  \begin{subfigure}{0.48\textwidth}
    \centering
    \includegraphics[width=\linewidth]{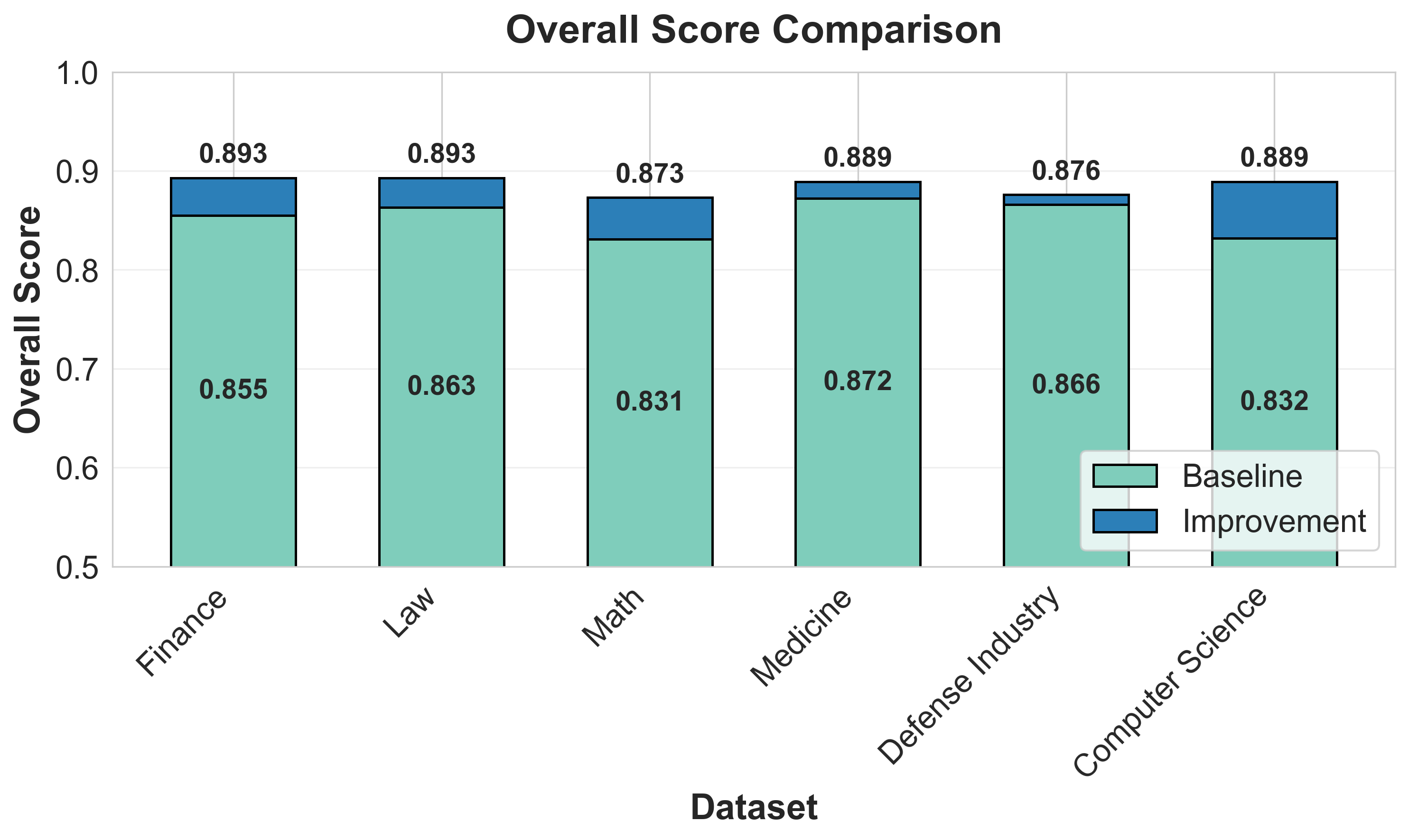}
    \caption{Overall performance comparison.}
    \label{fig:overall}
  \end{subfigure}\hfill
  \begin{subfigure}{0.48\textwidth}
    \centering
    \includegraphics[width=\linewidth]{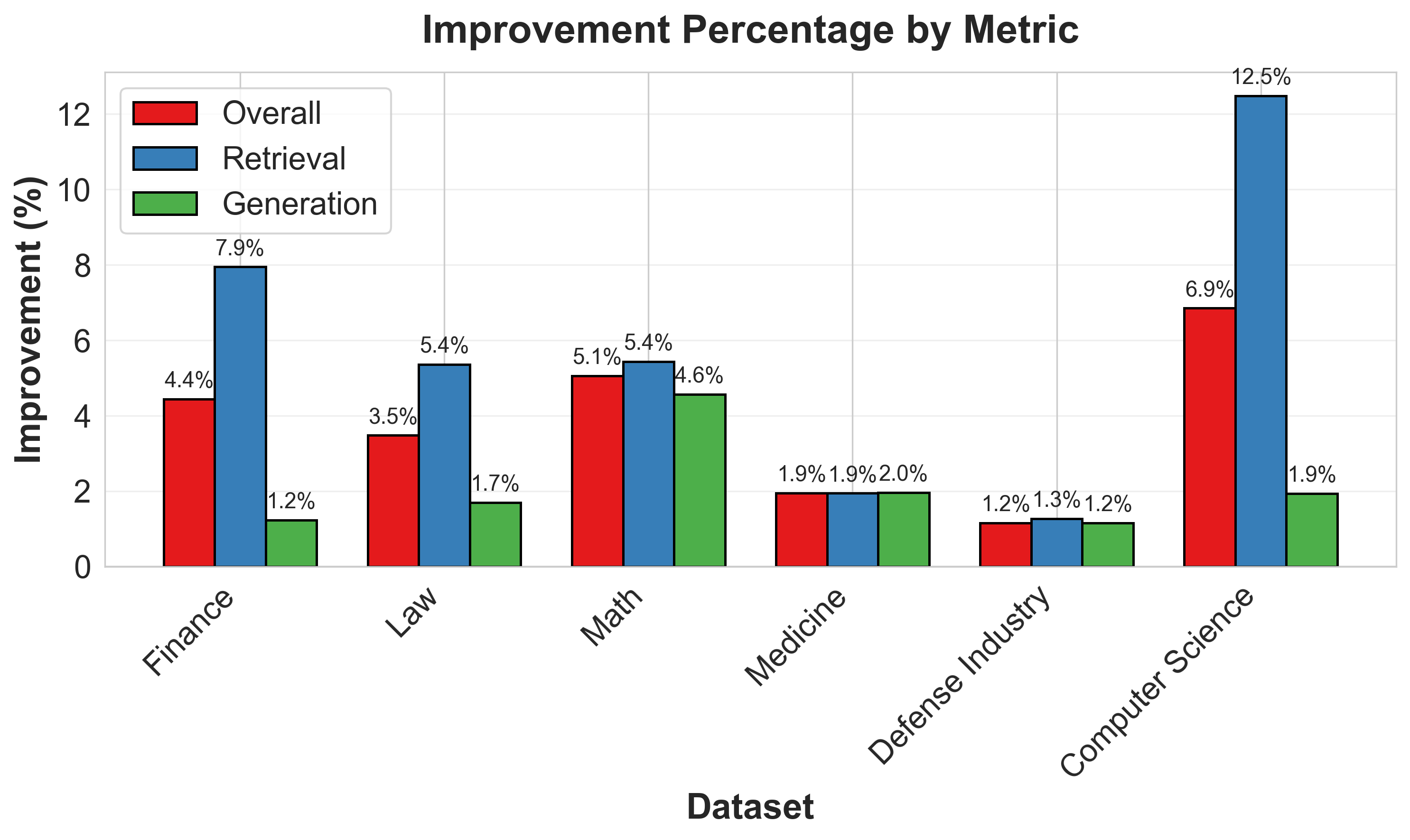}
    \caption{Relative improvement percentages.}
    \label{fig:improvements}
  \end{subfigure}
  \caption{Overall performance and improvement percentages obtained by \method.}
  \label{fig:overall_improvements}
\end{figure}

Applying \method~to the six distinct datasets, we obtained substantial improvements over the naive RAG baseline. As illustrated in \autoref{fig:retrieval_generation}, the highest improvement in retrieval performance was observed in the Computer Science dataset (+12.5\%), whereas the greatest improvement in generation quality was found in the Mathematics dataset. The overall highest combined score (mean of retrieval and generation metrics) was achieved in Computer Science (+6.9\% overall), followed by Mathematics (+5.1\%), as shown in \autoref{fig:overall_improvements}.

Table~\ref{tab:overall_performance} presents the overall performance comparison between our optimized RAG configurations and the naive RAG baseline across all datasets.

\begin{table}[!t]
\centering
\caption{Overall Performance: Best Configuration vs. Naive RAG Baseline}
\label{tab:overall_performance}
\begin{tabular}{lccc}
\toprule
\textbf{Dataset} & \textbf{Best Overall} & \textbf{Baseline Overall} & \textbf{Improvement} \\
\midrule
Finance & 0.893 & 0.855 & +4.4\% \\
Law & 0.893 & 0.863 & +3.5\% \\
Mathematics & 0.873 & 0.831 & +5.1\% \\
Medicine & 0.889 & 0.872 & +1.9\% \\
Defense Industry & 0.876 & 0.866 & +1.2\% \\
Computer Science & 0.889 & 0.832 & +6.9\% \\
\midrule
\textbf{Average} & \textbf{0.886} & \textbf{0.853} & \textbf{+3.8\%} \\
\bottomrule
\end{tabular}
\end{table}

\paragraph{Improvements} The results demonstrate consistent improvements across all domains, with the most substantial gains observed in Computer Science (+6.9\%) and Mathematics (+5.1\%), while more modest improvements were achieved in Defense Industry (+1.2\%) and Medicine (+1.9\%). These variations in improvement magnitude correlate strongly with dataset characteristics and question type distributions.

A detailed comparison of retrieval and generation metrics between the best-performing RAG combinations discovered by \method~and the naive baseline is provided in \autoref{tab:results_overview} and \autoref{tab:llm_semantic_overall}. These results demonstrate that automated evolutionary search can effectively discover configurations that substantially outperform standard pipeline setups across diverse knowledge domains.

\begin{table}[!t]
\centering
\small
\setlength{\tabcolsep}{8pt}
\caption{Comparison of Best vs.\ Baseline results across domains for Retrieval metrics. Overall is the mean of Recall@5, mAP, nDCG@5, and MRR (higher is better). Rows labeled \%$\Delta$ show relative change of Best over Baseline.}
\label{tab:results_overview}
\begin{tabular}{llccccc}
\toprule
\multirow{2}{*}{Dataset} & \multirow{2}{*}{Method} & \multirow{2}{*}{Overall$\uparrow$} & \multirow{2}{*}{Recall@5$\uparrow$} & \multirow{2}{*}{mAP$\uparrow$} & \multirow{2}{*}{nDCG@5$\uparrow$} & \multirow{2}{*}{MRR$\uparrow$} \\
 &  &  &  &  &  &  \\
\midrule
Finance & Best & 0.883 & 0.895 & 0.813 & 0.871 & 0.952 \\ 
        & Baseline & 0.819 & 0.848 & 0.764 & 0.800 & 0.862 \\
        & \%$\Delta$ & $+7.8\%$ & $+5.5\%$ & $+6.4\%$ & $+8.9\%$ & $+10.4\%$ \\
\midrule
Law     & Best & 0.885 & 0.888 & 0.828 & 0.875 & 0.948 \\ 
        & Baseline & 0.840 & 0.875 & 0.782 & 0.825 & 0.877 \\
        & \%$\Delta$ & $+5.4\%$ & $+1.5\%$ & $+5.9\%$ & $+6.1\%$ & $+8.1\%$ \\
\midrule
Math    & Best & 0.852 & 0.893 & 0.781 & 0.844 & 0.892 \\ 
        & Baseline & 0.808 & 0.845 & 0.753 & 0.792 & 0.844 \\
        & \%$\Delta$ & $+5.4\%$ & $+5.7\%$ & $+3.7\%$ & $+6.6\%$ & $+5.7\%$ \\
\midrule
Medicine & Best & 0.885 & 0.888 & 0.828 & 0.875 & 0.948 \\ 
         & Baseline & 0.876 & 0.906 & 0.824 & 0.860 & 0.914 \\
         & \%$\Delta$ & $+1.0\%$ & $-2.0\%$ & $+0.5\%$ & $+1.7\%$ & $+3.7\%$ \\
\midrule
Defense Industry & Best & 0.878 & 0.904 & 0.810 & 0.867 & 0.930 \\ 
                 & Baseline & 0.867 & 0.881 & 0.818 & 0.851 & 0.919 \\
                 & \%$\Delta$ & $+1.3\%$ & $+2.6\%$ & $-1.0\%$ & $+1.9\%$ & $+1.2\%$ \\
\midrule
Computer Science & Best & 0.883 & 0.890 & 0.823 & 0.873 & 0.945 \\ 
                 & Baseline & 0.785 & 0.837 & 0.721 & 0.763 & 0.818 \\
                 & \%$\Delta$ & $+12.5\%$ & $+6.3\%$ & $+14.1\%$ & $+14.4\%$ & $+15.5\%$ \\
\bottomrule
\end{tabular}
\end{table}

\begin{table}[!t]
\centering
\setlength{\tabcolsep}{4pt}
\caption{Comparison of Best vs.\ Baseline results across domains for LLM-based and Semantic evaluation metrics. Overall is the mean of the two for each dataset (higher is better). Additional \%$\Delta$ columns show relative change of Best over Baseline.}
\label{tab:llm_semantic_overall}
\resizebox{\linewidth}{!}{%
\begin{tabular}{lccccccccc}
\toprule
\multirow{2}{*}{Dataset} & \multicolumn{3}{c}{Overall $\uparrow$} & \multicolumn{3}{c}{LLM $\uparrow$} & \multicolumn{3}{c}{Semantic $\uparrow$} \\
\cmidrule(lr){2-4}\cmidrule(lr){5-7}\cmidrule(lr){8-10}
& Best & Baseline & \%$\Delta$ & Best & Baseline & \%$\Delta$ & Best & Baseline & \%$\Delta$ \\
\midrule
Finance           & 0.903 & 0.893 & $+1.1\%$ & 0.890 & 0.870 & $+2.3\%$ & 0.916 & 0.915 & $+0.1\%$ \\
Law               & 0.908 & 0.887 & $+2.4\%$ & 0.889 & 0.857 & $+3.7\%$ & 0.927 & 0.916 & $+1.2\%$ \\
Math              & 0.893 & 0.855 & $+4.4\%$ & 0.871 & 0.810 & $+7.5\%$ & 0.916 & 0.899 & $+1.9\%$ \\
Medicine          & 0.903 & 0.867 & $+4.2\%$ & 0.881 & 0.837 & $+5.3\%$ & 0.924 & 0.898 & $+2.9\%$ \\
Defense Industry  & 0.875 & 0.865 & $+1.2\%$ & 0.837 & 0.823 & $+1.7\%$ & 0.913 & 0.906 & $+0.8\%$ \\
Computer Science  & 0.895 & 0.879 & $+1.8\%$ & 0.887 & 0.860 & $+3.1\%$ & 0.903 & 0.897 & $+0.7\%$ \\
\bottomrule
\end{tabular}
}%
\end{table}

\subsection{Retrieval Performance Analysis}

Computer Science demonstrates the most dramatic retrieval improvement (+12.5\%), particularly in mAP (+14.1\%) and nDCG@5 (+14.4\%). This can be attributed to its lower chunk density (34.4 chunks/article) making the multi-query expansion particularly effective—fewer chunks mean that semantic diversification through query expansion can more effectively cover the search space. Wikipedia Computer Science articles tend to be concise and definition-focused (e.g., articles on specific algorithms, data structures, or programming concepts), with less narrative elaboration compared to other domains. The combination of \texttt{multi\_query} with \texttt{vector\_retrieval} proved especially beneficial because technical terminology can be expressed in multiple ways (e.g., "hash table" vs. "hash map" vs. "dictionary structure"), and query expansion helps capture these lexical variations.

Medicine, despite having the highest baseline retrieval performance (R@5: 0.906), shows more modest improvements (+1.9\%), suggesting that the naive RAG baseline was already performing reasonably well. Wikipedia medical articles are typically comprehensive and well-structured, following consistent organizational patterns (e.g., symptoms, causes, diagnosis, treatment) with standardized medical terminology. This high information density (50.9 chunks/article) and terminological consistency means that even simple vector retrieval performs well. The high chunk density necessitated sophisticated reranking strategies (\texttt{hybrid\_rerank}) to efficiently process large candidate sets and identify the most relevant passages among many potentially relevant sections.

\subsection{Generation Performance Analysis}

The generation metrics reveal that improvements are primarily driven by enhanced LLM scoring rather than semantic similarity, as shown in Table~\ref{tab:llm_semantic_overall}. Mathematics demonstrates the most significant LLM score improvement (+7.5\%), which directly correlates with its high proportion of \textit{long-answer} questions (41\%). The combination of \texttt{contextual\_chunk\_headers} and \texttt{long\_context\_reorder} prompt maker proved particularly effective for mathematical encyclopedia articles. Wikipedia mathematics articles often explain concepts through hierarchical relationships (e.g., how a theorem relates to lemmas, corollaries, and applications), and contextual chunk headers help preserve these conceptual dependencies when chunks are retrieved separately. This is crucial for generating coherent explanations that connect abstract definitions to their properties and applications.

This pattern is reinforced by Computer Science's strong LLM score gains (+3.1\%), which also benefits from contextual augmentation through \texttt{prev\_next\_augmenter}. Wikipedia Computer Science articles often explain concepts sequentially (e.g., describing an algorithm's steps, or explaining how data structures work through examples), making contiguous context particularly valuable for maintaining logical flow in generated explanations.

Conversely, datasets with high \textit{interpretation} question ratios (Medicine: 53\%, Finance: 51\%, Law: 49\%) show smaller LLM score improvements (+5.3\%, +2.3\%, +3.7\% respectively), suggesting that current RAG optimization techniques are less effective for questions requiring inferential reasoning beyond direct information retrieval.

\subsection{Retrieval Module Analysis}

The retrieval phase in our modular pipeline comprises five stages: \textit{Pre-Embedding, Query Expansion, Retrieval, Reranking,} and \textit{Passage Filtering}. Their optimal selections for each dataset are shown in \autoref{tab:best_retrieval_combinations}.

\begin{table*}[!t]
\centering
\small
\caption{Best-performing \textbf{Retrieval-side} module combinations discovered by \method~for each dataset. The retrieval pipeline includes Pre-Embedding, Query Expansion, Retrieval, Reranking, and Passage Filtering. "--" indicates the module was not applied in that configuration.}
\label{tab:best_retrieval_combinations}
\resizebox{\textwidth}{!}{
\begin{tabular}{lccccc}
\toprule
\textbf{Dataset} & \textbf{Pre-Embedding} & \textbf{Query Expansion} & \textbf{Retrieval} & \textbf{Rerank} & \textbf{Passage Filter} \\
\midrule
Finance & -- & multi\_query & vector\_retrieval & cross\_encoder & similarity\_threshold \\
Law & -- & multi\_query & vector\_retrieval & cross\_encoder & similarity\_threshold \\
Mathematics & contextual\_chunk\_headers & multi\_query & vector\_retrieval & hybrid\_rerank & similarity\_threshold \\
Medicine & -- & -- & vector\_retrieval & hybrid\_rerank & similarity\_threshold \\
Defense Industry & -- & -- & vector\_retrieval & hybrid\_rerank & simple\_threshold \\
Computer Science & -- & multi\_query & vector\_retrieval & hybrid\_rerank & similarity\_threshold \\
\bottomrule
\end{tabular}
}
\end{table*}

Across all domains, \texttt{vector\_retrieval} emerged as the dominant retriever. This consistency reflects its balanced performance between semantic precision and contextual coherence, suggesting that its embedding granularity generalizes effectively across domains with varying lexical overlap and content densities.

The prevalence of \texttt{multi-query expansion} in four out of six datasets (Finance, Law, Mathematics, Computer Science) confirms its role in enhancing recall through semantic diversification. Notably, Medicine and Defense Industry omit query expansion, which correlates with their high chunk densities (50.9 and 50.6 chunks/article respectively)—with more chunks already available, exhaustive query expansion may introduce diminishing returns or noise. Other query expansion modules such as \texttt{decomposition} and \texttt{step-back prompting} were consistently pruned during evolution, suggesting they suffer from overly restrictive assumptions about query structure that limit their generalizability.

The \texttt{pre-embedding} stage, utilized only in Mathematics through \texttt{contextual\_chunk\_headers}, proved advantageous for this domain where Wikipedia articles are organized hierarchically. Mathematical encyclopedia entries typically structure content from general concepts to specific applications (e.g., a calculus article moving from definitions to theorems to examples), and enriching chunks with their section context helps maintain these conceptual relationships during retrieval. This aligns with Mathematics having the highest proportion of \textit{long-answer} questions (41\%), which benefit from understanding how different parts of an article relate to each other.

Reranking proved essential across all domains, with \texttt{cross\_encoder} selected for lower-density datasets (Finance: 43.5, Law: 37.3 chunks/article) and \texttt{hybrid\_rerank} (hybrid reranker) for higher-density datasets (Medicine: 50.9, Defense Industry: 50.6 chunks/article). This pattern reflects a trade-off: cross-encoders provide more thorough evaluation suitable when candidate sets are smaller, while parallel hybrid reranking maintains competitive performance when processing larger candidate sets.

\subsection{Generation Module Analysis}

Following retrieval, generation involves \textit{Passage Augmentation, Passage Compression, Prompt Making,} and \textit{Post-Generation} phases. Their corresponding best-performing modules are listed in \autoref{tab:best_generation_combinations}.

\begin{table*}[!t]
\centering
\small
\caption{Best-performing \textbf{Generation-side} module combinations discovered by \method~for each dataset. The generation pipeline includes Passage Augmentation, Passage Compression, Prompt Making, and Post-Generation phases. "--" indicates the module was not applied in that configuration.}
\label{tab:best_generation_combinations}
\resizebox{\textwidth}{!}{
\begin{tabular}{lcccc}
\toprule
\textbf{Dataset} & \textbf{Passage Augment} & \textbf{Passage Compression} & \textbf{Prompt Maker} & \textbf{Post-Generation} \\
\midrule
Finance & prev\_next\_augmenter & -- & -- & reflection\_revising \\
Law & relevant\_segment\_extractor & -- & long\_context\_reorder & reflection\_revising \\
Mathematics & prev\_next\_augmenter & -- & long\_context\_reorder & reflection\_revising \\
Medicine & prev\_next\_augmenter & -- & long\_context\_reorder & reflection\_revising \\
Defense Industry & relevant\_segment\_extractor & -- & long\_context\_reorder & reflection\_revising \\
Computer Science & prev\_next\_augmenter & -- & -- & reflection\_revising \\
\bottomrule
\end{tabular}
}
\end{table*}

The selection of \texttt{reflection\_revising} across all domains highlights its general utility in improving factual consistency and narrative coherence through iterative self-refinement. This post-generation module acts as a dynamic correction mechanism, proving particularly effective for \textit{long-answer} questions.

Notably, \texttt{passage compression} was never selected by the evolutionary search, implying that compression-induced information loss outweighed any gain in input efficiency. This finding is significant: it demonstrates that the genetic algorithm correctly learned to prune ineffective design branches, and suggests that for current LLM context window sizes, information enrichment strategies are more valuable than reduction techniques.

The choice between \texttt{prev\_next\_augmenter} (4 datasets) and \texttt{relevant\_segment\_extractor} (2 datasets) reflects different needs for contextual window flexibility. Fixed-window augmentation (\texttt{prev\_next\_augmenter}) dominates in domains where Wikipedia articles have relatively uniform information density and consistent chunk relevance patterns: Mathematics (where related concepts are typically adjacent), Medicine (standardized article structures), Computer Science (sequential explanations), and Finance (connected economic principles). Conversely, adaptive-window extraction (\texttt{relevant\_segment\_extractor}) proves beneficial in Law and Defense Industry, where Wikipedia articles exhibit variable information density. Legal articles may have lengthy historical sections followed by brief but crucial legal provisions, while Defense Industry articles might have extensive operational history but concise technical specifications. The adaptive approach allows the system to dynamically adjust the context window size based on where relevant information is concentrated, rather than using a fixed window that might include irrelevant sections or miss relevant distant context.

The \texttt{long\_context\_reorder} prompt maker appears in four datasets (Law, Mathematics, Medicine, Defense Industry), with a clear pattern: it is selected for datasets with either high chunk density (Medicine: 50.9, Defense: 50.6 chunks/article) or high average tokens per chunk (Law: 182.2, Medicine: 190.3 tokens/chunk). This reordering strategy helps manage the "lost in the middle" problem for long context windows, ensuring relevant information is positioned optimally for the LLM's attention mechanism.

\subsection{Dataset-Specific Analysis and Insights}

\paragraph{Computer Science (+6.9\% overall, +12.5\% retrieval).}
The Computer Science dataset achieved the largest improvements, driven by a synergistic combination of low chunk density (34.4 chunks/article, lowest among all datasets) and balanced question distribution (30\% \textit{factual}, 37\% \textit{interpretation}, 33\% \textit{long-answer}). Wikipedia Computer Science articles are typically concise and focused on specific concepts (e.g., individual algorithms, data structures, or programming paradigms), resulting in fewer but more focused chunks per article. This low chunk density makes \texttt{multi-query expansion} particularly effective, as semantic diversification can thoroughly cover the compact search space without introducing excessive noise. The combination with \texttt{hybrid\_rerank} and \texttt{prev\_next\_augmenter} proves especially effective because Computer Science encyclopedia entries often explain concepts through sequential progression (defining a concept, explaining how it works, then providing examples). The dramatic mAP improvement (+14.1\%) indicates that the optimized pipeline significantly improves ranking quality, moving the most relevant conceptual explanations to top positions.

\paragraph{Mathematics (+5.1\% overall, +5.4\% retrieval).}

Mathematics dataset uniquely employs \texttt{contextual\_chunk\_headers}, a choice directly motivated by its highest proportion of \textit{long-answer} questions (41\%). Wikipedia mathematics articles typically organize content hierarchically, with main topics subdivided into definitions, properties, examples, and applications. When these sections are chunked, the semantic relationship between them (e.g., that an "Example" section refers to concepts defined earlier) can be lost. Contextual chunk headers address this by enriching each chunk with its section context, helping the retrieval system understand that a chunk about "Applications of Fourier Transform" is related to chunks about "Fourier Transform." The synergy between \texttt{contextual\_chunk\_headers}, \texttt{multi-query expansion}, and \texttt{long\_context\_reorder} creates a pipeline optimized for retrieving and organizing related mathematical concepts. The substantial LLM score improvement (+7.5\%) validates that these techniques help the model generate coherent explanations that properly connect definitions to their properties and applications.

\paragraph{Law (+3.5\% overall, +5.4\% retrieval) and Finance (+4.4\% overall, +7.9\% retrieval).}
Both domains exhibit high \textit{interpretation} question ratios (Law: 49\%, Finance: 51\%), which limits the effectiveness of retrieval-focused optimizations. However, moderate improvements are achieved through precision-focused strategies. Wikipedia legal and financial articles tend to use domain-specific terminology consistently, making terminology matching relatively straightforward for baseline retrieval. The combination of \texttt{multi-query expansion}, \texttt{vector\_retrieval}, and \texttt{cross\_encoder reranking} improves precision, with moderate chunk densities (Law: 37.3, Finance: 43.5 chunks/article) permitting thorough cross-encoder evaluation.

Law uniquely employs \texttt{relevant\_segment\_extractor}, reflecting the variable information density in Wikipedia legal articles. These articles often contain sections of vastly different lengths and relevance densities—for instance, a lengthy historical background section might be followed by a brief but information-dense section on legal provisions, or vice versa. The adaptive context window allows the system to expand or contract the retrieved segment based on where the relevant information is concentrated, rather than using a fixed window that might include excessive irrelevant historical context or miss relevant adjacent legal interpretations. The \texttt{long\_context\_reorder} prompt maker helps manage relatively lengthy chunks (182.2 tokens/chunk average) by positioning the most relevant segments optimally.

Finance's configuration uses \texttt{prev\_next\_augmenter}, reflecting that Wikipedia finance articles often explain concepts through connected principles (e.g., explaining how interest rates affect various economic indicators in sequence). The modest semantic similarity improvement (+0.1\%) suggests that financial terminology is already well-captured by baseline embeddings, with gains primarily from improved LLM reasoning about relationships between concepts (+2.3\% LLM score).

\paragraph{Medicine (+1.9\% overall) and Defense Industry (+1.2\% overall).}
Both datasets display the smallest improvements, despite different underlying reasons. Medicine's high baseline performance (0.872) and highest \textit{interpretation} ratio (53\%) suggest the naive RAG already performs well. Wikipedia medical articles follow highly standardized structures (symptoms, causes, diagnosis, treatment, epidemiology) with consistent medical terminology, making even simple retrieval effective. The highest chunk density (50.9 chunks/article) and token count (106,552 total tokens) reflect the encyclopedic comprehensiveness of medical articles, providing abundant information where most chunks are potentially relevant. This reduces the marginal benefit of sophisticated retrieval optimization. The configuration's omission of query expansion reinforces this: medical terminology is sufficiently standardized that single-query retrieval suffices.

Defense Industry's modest improvement (+1.2\%) correlates with its high \textit{interpretation} ratio (46\%) and high chunk density (50.6 chunks/article). Wikipedia military and defense articles are often comprehensive, covering historical development, technical specifications, operational use, and variants within single articles, with highly variable section lengths. The \texttt{relevant\_segment\_extractor} proves valuable here by adaptively determining context window size—for instance, expanding to include multiple chunks when technical specifications are spread across several sections, or contracting when a single chunk contains concentrated relevant information.

\subsection{Question Type Impact on Technique Effectiveness}

 Our analysis reveals a strong inverse correlation between \textit{interpretation} question ratio and overall improvement magnitude. Datasets with \textit{interpretation} ratios above 45\% (Medicine: 53\%, Finance: 51\%, Law: 49\%, Defense: 46\%) show average improvement of +2.8\%, while those below 40\% (Computer Science: 37\%, Mathematics: 34\%) show average improvement of +6.0\%.

This pattern suggests that current RAG optimization techniques are more effective for \textit{factual} and \textit{long-answer} questions, which benefit directly from improved retrieval precision and contextual augmentation. \textit{interpretation} questions, requiring nuanced understanding of implications and contextual inference beyond retrieved text, are less amenable to retrieval-focused optimizations. These questions likely require reasoning capabilities that current LLMs either possess or lack largely independent of retrieval quality, explaining why even sophisticated RAG configurations show limited improvement.

\textit{long-answer} questions exhibit the strongest positive correlation with contextual techniques. Mathematics and Computer Science, with \textit{long-answer} ratios of 41\% and 33\% respectively, show the largest LLM score improvements (+7.5\% and +3.1\%) when employing \texttt{prev\_next\_augmenter} or \texttt{contextual\_chunk\_headers}. These techniques help maintain coherence when generating comprehensive encyclopedia-style explanations that integrate information from multiple sections of an article (e.g., combining a definition with its properties, examples, and applications into a single coherent answer).

\subsection{Cross-Domain Patterns and Genetic Search Dynamics}

Across all six datasets, several modules appeared recurrently among top configurations: \texttt{vector\_retrieval} for retrieval (6/6 datasets) and \texttt{reflection\_revising} for post-generation (6/6 datasets). This consistency indicates that these components form a "robust RAG backbone"—a minimal set of effective building blocks that generalize well across domains with varying semantic density, question types, and content structures.

The genetic search proved to be an effective strategy for exploring the configuration space without requiring exhaustive evaluation. Across all datasets, the algorithm consistently identified high-performing configurations after a limited number of evaluations, demonstrating its capability to efficiently navigate the large and diverse search space. By maintaining a balanced interplay between exploration of new combinations and exploitation of promising patterns, the search process produced robust results within a reasonable computational budget, confirming the practicality of evolutionary optimization for modular RAG pipelines.

The systematic absence of certain modules provides valuable insights. \texttt{Passage compression} was never selected, indicating the algorithm correctly learned that compression-induced information loss outweighs efficiency gains. Alternative query expansion methods (\texttt{decomposition}, \texttt{step-back prompting}) were pruned in favor of \texttt{multi-query expansion}, suggesting that flexible, unrestricted query generation generalizes better than structured approaches making assumptions about query format.

\subsection{Broader Implications and Design Principles}

The observed results reinforce several key principles for effective RAG pipeline design:

\paragraph{Robust Components.} The consistent selection of \texttt{vector\_retrieval} for retrieval and \texttt{reflection\_revising} for post-generation across all domains suggests these should be considered foundational components. Their effectiveness indicates they provide fundamental capabilities that benefit diverse question types and content structures.

\paragraph{Domain-Adaptive Augmentation.} The choice between fixed-window augmentation (\texttt{prev\_next\_augmenter}) and adaptive-window extraction (\texttt{relevant\_segment\_extractor}) should be guided by information density uniformity: fixed windows work well for domains with consistent relevance patterns and uniform section lengths; adaptive windows excel when articles have highly variable section lengths and information density, allowing dynamic adjustment to avoid including irrelevant context or missing relevant distant information.

\paragraph{Density-Aware Reranking.} Dataset chunk density may inform reranker selection: cross-encoders for thorough evaluation when candidate sets are small (<40 chunks/article); hybrid rerankers (cross-encoder + LLM) for computational efficiency when candidate sets are large (>50 chunks/article). These tendencies should be read qualitatively. Their strength varies with hyperparameters (k, candidate pool size), and context budget; a fuller ablation is left to future work.

\paragraph{Question-Type Awareness.} Pipeline optimization effectiveness varies significantly by question type distribution. Datasets with higher proportions of \textit{interpretation} questions (Medicine: 53\%, Finance: 51\%, Law: 49\%) show smaller improvements (+1.2\% to +4.4\%) compared to those with more \textit{factual} and \textit{long-answer} questions (Computer Science: +6.9\%, Mathematics: +5.1\%). This suggests that current RAG optimization techniques primarily improve information retrieval and contextual organization, with limited impact on inferential reasoning capabilities.

\subsection{Limitations and Future Directions}

While our results demonstrate consistent improvements across diverse domains, several limitations warrant discussion:

\paragraph{\textit{interpretation} Question Challenge.} The modest improvements on \textit{interpretation}-heavy datasets suggest that current RAG techniques are less effective for questions requiring deep contextual understanding and inferential reasoning. Future work should explore techniques specifically designed for interpretive tasks, such as multi-hop reasoning chains, or explicit reasoning trace generation.

\paragraph{Computational Trade-offs.} Advanced techniques like parallel reranking and reflection-based revision introduce significant computational costs. While our fitness function weights performance heavily, production deployments must consider latency and resource constraints. Future work could extend the framework toward multi-objective optimization, incorporating latency and cost-awareness alongside retrieval and generation quality.

\paragraph{Limited Cross-Domain Generalization.} The optimal configurations vary substantially across domains (e.g., Medicine omits query expansion while Computer Science requires it), suggesting limited zero-shot generalization. Transfer learning approaches or meta-learning could enable more efficient configuration adaptation to new domains, potentially using dataset characteristics (chunk density, question distribution, token statistics) as features for predicting effective configurations.

\paragraph{Question Type Granularity.} Our three-category taxonomy (\textit{factual}, \textit{interpretation}, \textit{long-answer}) may be too coarse to capture nuances in question complexity. Finer-grained categorization (e.g., single-hop vs. multi-hop reasoning, comparison questions, causal reasoning) could reveal more targeted optimization opportunities.

\paragraph{Rapidly Advancing RAG Techniques.} \method~ presents a pipeline for analyzing RAG technique combinations and finding suitable combinations for given datasets. However, RAG is a rapidly advancing field and new advanced RAG techniques are developed constantly. Our pipeline proposes a framework and categorization which may welcome new techniques such as HyperGraph retrieval. Future work could extend \method~ with these new techniques to form a more comprehensive RAG technique analysis pipeline.

\subsection{Key Contributions and Findings}

\paragraph{Holistic Pipeline Optimization.} We cast RAG design as a joint, end-to-end configuration problem and optimize \emph{entire} pipelines rather than greedily selecting per-module “best” components. To tractably navigate the combinatorial space—where exhaustive grid search is impractical due to end-to-end runtime—we employ an evolutionary (genetic) search procedure that uncovers high-performing configurations under realistic compute budgets.

\paragraph{Question-Type Sensitivity Framework.} We quantify how the question-type mixture governs both achievable gains and where to invest optimization effort. Datasets dominated by \textit{long-answer}/\textit{factual} queries (e.g., Computer Science: 63\%, Mathematics: 66\%) realize larger improvements (up to +6.9\% and +5.1\%), while \textit{interpretation}-heavy domains (Medicine: 53\%, Finance: 51\%, Law: 49\%) yield smaller deltas (+1.9\%–+4.4\%). This highlights a persistent gap in current RAG methods for inference-heavy reasoning.

\paragraph{Domain-Specific Optimization Guidelines.} We provide actionable mappings from data characteristics to technique choices \emph{and} summarize the recurring design patterns surfaced by search. High chunk density/length favors cross-encoder or hybrid rerankers; non-uniform or hierarchical content benefits from adaptive windowing or segment extraction; and the question-type distribution dictates the balance between retrieval tuning and generation refinement. In practice, strong configurations repeatedly include contextual augmentation (\texttt{prev\_next\_augmenter} or \texttt{relevant\_segment\_extractor}) together with sophisticated reranking—offering a concrete recipe for practitioners.

\paragraph{Robust RAG Backbone.} Across various subjects, \texttt{vector\_retrieval} (retriever) paired with \texttt{reflection\_revising} (post-generation refinement) consistently anchors the best pipelines, forming a subject-agnostic backbone. Other modules—query expansion, reranking, passage augmentation, and prompt maker—serve as adaptive attachments selected by evolutionary search to match dataset-specific characteristics, enabling scalable multi-domain deployment.

%% file: sections/06-conclusion.tex
\section{Conclusion}

This work presents a comprehensive investigation into domain-specific optimization of modular RAG pipelines through evolutionary search. By applying \method~to six diverse Wikipedia-based datasets spanning Mathematics, Law, Finance, Medicine, Defense Industry, and Computer Science, we demonstrate that automated configuration discovery can consistently outperform naive RAG baselines, achieving an average improvement of +3.8\% overall (ranging from +1.2\% to +6.9\% across domains).

Our experimental results reveal several critical insights into RAG system design. First, we establish that question type distribution serves as a strong predictor of optimization potential: datasets dominated by \textit{factual} and \textit{long-answer} questions (Computer Science: 63\%, Mathematics: 66\%) achieve substantially larger improvements (+6.9\% and +5.1\%) compared to \textit{interpretation}-heavy datasets (Medicine: 53\%, Finance: 51\%, Law: 49\%) which show more modest gains (+1.9\% to +4.4\%). This pattern suggests that current RAG optimization techniques excel at improving information retrieval and contextual organization but have limited impact on enhancing inferential reasoning capabilities—a finding that highlights important directions for future research.

Second, we identify a "robust RAG backbone" consisting of \texttt{vector\_retrieval} retrieval and \texttt{reflection\_revising} post-generation, which appear across all optimal configurations. This consistency across diverse domains indicates that these components provide fundamental capabilities that generalize effectively. Beyond this backbone, domain-specific modules emerge based on dataset characteristics: chunk density determines reranker selection (cross-encoder for sparse datasets, hybrid reranker for dense ones); information density uniformity guides augmentation strategy choice (fixed-window for uniform content, adaptive-window for variable density); and the presence of contextual dependencies motivates pre-embedding enrichment strategies.

Third, the genetic search approach demonstrates remarkable efficiency, achieving high-quality configurations while evaluating less than 0.2\% of the 46,080 possible architectures. The algorithm's ability to systematically prune ineffective modules—such as passage compression (never selected) and overly restrictive query expansion methods—validates evolutionary optimization as a practical approach for navigating large RAG design spaces where exhaustive search or manual tuning is infeasible.

Our analysis of individual datasets reveals nuanced patterns in how Wikipedia article structure influences optimal RAG configuration. Computer Science's dramatic retrieval improvement (+12.5\%) stems from the synergy between low chunk density (34.4 chunks/article) and multi-query expansion's ability to cover the compact search space thoroughly. Mathematics' unique adoption of contextual chunk headers directly addresses the hierarchical organization of mathematical encyclopedia entries, enabling +7.5\% improvement in LLM-based generation scores by preserving relationships between definitions, properties, and applications. Conversely, Medicine's modest improvement (+1.9\%) reflects the domain's already-strong baseline performance, attributable to standardized article structures and consistent medical terminology that make even simple retrieval effective.

The work also exposes important limitations that warrant future investigation. The modest improvements on \textit{interpretation}-heavy datasets indicate a fundamental challenge: these questions require reasoning capabilities that current RAG techniques cannot adequately address through improved retrieval alone. Additionally, our approach identifies complete optimal configurations without isolating individual component contributions, leaving synergistic and antagonistic interactions between modules incompletely understood. Future work incorporating factorial ablation studies could elucidate these interactions and enable more targeted optimization strategies.

Looking forward, several promising research directions emerge from this work. First, the development of RAG techniques specifically designed for interpretive reasoning—potentially incorporating multi-hop reasoning chains, explicit reasoning traces, or knowledge graph integration—could address the performance gap on \textit{interpretation} questions. Second, multi-objective optimization incorporating computational cost and latency alongside performance metrics would enable practical deployment trade-offs. Third, meta-learning approaches leveraging dataset characteristics (chunk density, question distribution, content structure) could enable efficient zero-shot configuration prediction for new domains, reducing the need for domain-specific optimization.

A critical methodological contribution of this work lies in optimizing complete pipeline configurations rather than independently selecting "best" modules from each category. Traditional RAG optimization approaches evaluate techniques in isolation—identifying the best retrieval method, best reranking strategy, and best generation approach separately—then combine these individual "winners" into a final pipeline. However, this greedy, per-module optimization ignores potential synergies and conflicts between components. Our evolutionary approach, by contrast, evaluates entire configurations holistically, allowing modules that may be suboptimal in isolation to emerge when they enhance the performance of other pipeline components. This configuration-level optimization is essential: a reranker that excels with one retrieval method may perform poorly with another, or an augmentation strategy that benefits one generation approach may hinder another. The systematic absence of certain modules despite their individual strengths (e.g., passage compression, decomposition-based query expansion) demonstrates that \method~correctly identifies when components that work well in isolation degrade overall pipeline performance.

The modular framework and evolutionary search methodology presented here provide both a practical system for RAG optimization and a research platform for investigating how different pipeline components interact across diverse domains. As the RAG landscape continues to evolve with emerging techniques such as HyperGraph retrieval and advanced reasoning modules, \method's extensible architecture offers a principled approach to evaluating and integrating new capabilities. By establishing that optimal RAG configurations are fundamentally domain-dependent—with improvement magnitudes varying threefold across our datasets (1.2\% to 6.9\%)—this work underscores a crucial insight: there is no universal RAG solution, and thoughtful, data-driven optimization tailored to dataset characteristics and question types is essential for maximizing performance in production deployments.

In summary, this work contributes: (1) a holistic pipeline optimization approach that accounts for inter-component interactions rather than greedy per-module selection; (2) a question type sensitivity framework linking question distributions to optimization potential of RAG pipelines; (3) empirically-grounded design principles mapping dataset characteristics to effective module combinations; (4) identification of robust RAG components alongside domain-adaptive extensions. These contributions collectively advance our understanding of RAG system design and provide practitioners with both methodology and guidance for optimizing RAG pipelines across diverse domains and question types.

%% file: sections/07-acknowledgement.tex
\section{Acknowledgement}
We acknowledge the open-source RAG community whose tools and code informed this work, and the Wikipedia volunteer editors and the Wikimedia Foundation for making key data openly available. We used AI-assisted tools for copy-editing under the authors’ supervision; the authors remain responsible for all content and any errors.

%% file: main.bbl
\begin{thebibliography}{100}

\bibitem{lewis2020retrieval}
P.~Lewis, E.~Perez, A.~Piktus, F.~Petroni, V.~Karpukhin, N.~Goyal, H.~K\"{u}ttler, M.~Lewis, W.-t. Yih, T.~Rockt\"{a}schel, S.~Riedel, and D.~Kiela, ``Retrieval-augmented generation for knowledge-intensive nlp tasks,'' in {\em Advances in Neural Information Processing Systems}, 2020.

\bibitem{wang2023query}
L.~Wang, N.~Yang, X.~Huang, L.~Yang, R.~Majumder, and F.~Wei, ``Query2doc: Query expansion with large language models,'' {\em arXiv preprint arXiv:2303.07678}, 2023.

\bibitem{lin2024reranking}
W.~Lin, Z.~Qian, G.~Jiang, J.~Wang, Y.~Wu, X.~Wang, and M.~Zhang, ``Rankgpt: Is chatgpt good at search and reranking?,'' in {\em Proceedings of the 2024 Conference on Empirical Methods in Natural Language Processing}, pp.~9821--9836, 2024.

\bibitem{li2024contextual}
O.~Li, M.~Alikhani, and S.~Gehrmann, ``Contextual position encoding: Learning to count what's important,'' {\em arXiv preprint arXiv:2405.18719}, 2024.

\bibitem{shinn2024reflexion}
N.~Shinn, F.~Cassano, A.~Gopinath, K.~Narasimhan, and S.~Yao, ``Reflexion: Language agents with verbal reinforcement learning,'' {\em Advances in Neural Information Processing Systems}, vol.~36, 2024.

\bibitem{gao2023survey}
L.~Gao, Z.~Dai, and J.~Callan, ``Retrieval-augmented generation for large language models: A survey,'' {\em arXiv preprint arXiv:2312.10997}, 2023.

\bibitem{lewis2020rag}
P.~Lewis, E.~Perez, A.~Piktus, F.~Petroni, V.~Karpukhin, N.~Goyal, H.~K{\"u}ttler, M.~Lewis, W.-t. Yih, T.~Rockt{\"a}schel, S.~Riedel, and D.~Kiela, ``Retrieval-augmented generation for knowledge-intensive nlp tasks,'' in {\em Advances in Neural Information Processing Systems}, 2020.

\bibitem{karpukhin2020dpr}
V.~Karpukhin, B.~Oguz, S.~Min, P.~Lewis, L.~Wu, S.~Edunov, D.~Chen, and W.-t. Yih, ``Dense passage retrieval for open-domain question answering,'' in {\em Proceedings of the 2020 Conference on Empirical Methods in Natural Language Processing (EMNLP)}, 2020.

\bibitem{izacard2021fid}
G.~Izacard and E.~Grave, ``Leveraging passage retrieval with generative models for open domain question answering,'' in {\em Proceedings of the 16th Conference of the European Chapter of the Association for Computational Linguistics (EACL)}, 2021.
\newblock Also known as Fusion-in-Decoder (FiD).

\bibitem{chen2017drqa}
D.~Chen, A.~Fisch, J.~Weston, and A.~Bordes, ``Reading wikipedia to answer open-domain questions,'' in {\em Proceedings of the 55th Annual Meeting of the Association for Computational Linguistics (ACL)}, 2017.

\bibitem{guu2020realm}
K.~Guu, K.~Lee, Z.~Tung, P.~Pasupat, and M.-W. Chang, ``{REALM}: Retrieval-augmented language model pre-training,'' in {\em Proceedings of the 37th International Conference on Machine Learning (ICML)}, 2020.

\bibitem{nogueira2019passage}
R.~Nogueira and K.~Cho, ``Passage re-ranking with {BERT},'' in {\em Proceedings of the 2019 Conference of the North American Chapter of the Association for Computational Linguistics (NAACL) Deep Learning Approaches for Low-Resource NLP Workshop}, 2019.

\bibitem{luan2021sparse}
Y.~Luan, J.~Eisenstein, K.~Toutanova, and M.~Collins, ``Sparse, dense, and attentional representations for text retrieval,'' in {\em Transactions of the Association for Computational Linguistics (TACL)}, 2021.

\bibitem{johnson2017billion}
J.~Johnson, M.~Douze, and H.~J{\'e}gou, ``Billion-scale similarity search with {GPU}s,'' in {\em Proceedings of the IEEE Conference on Computer Vision and Pattern Recognition (CVPR)}, 2017.

\bibitem{malkov2018efficient}
Y.~A. Malkov and D.~A. Yashunin, ``Efficient and robust approximate nearest neighbor search using hierarchical navigable small world graphs,'' {\em IEEE Transactions on Pattern Analysis and Machine Intelligence}, vol.~40, no.~4, pp.~824--836, 2018.

\bibitem{reimers2019sentencebert}
N.~Reimers and I.~Gurevych, ``Sentence-{BERT}: Sentence embeddings using {S}iamese {BERT}-networks,'' in {\em Proceedings of the 2019 Conference on Empirical Methods in Natural Language Processing (EMNLP)}, 2019.

\bibitem{khattab2020colbert}
O.~Khattab and M.~Zaharia, ``{C}ol{BERT}: Efficient and effective passage search via contextualized late interaction over {BERT},'' in {\em Proceedings of the 43rd International ACM SIGIR Conference on Research and Development in Information Retrieval (SIGIR)}, 2020.

\bibitem{rocchio1971relevance}
J.~J. Rocchio, ``Relevance feedback in information retrieval,'' in {\em The SMART Retrieval System: Experiments in Automatic Document Processing}, 1971.

\bibitem{robertson2009probabilistic}
S.~Robertson and H.~Zaragoza, ``The probabilistic relevance framework: {BM25} and beyond,'' {\em Foundations and Trends in Information Retrieval}, vol.~3, no.~4, pp.~333--389, 2009.

\bibitem{nogueira2019doc2query}
R.~Nogueira, W.~Yang, J.~Lin, and K.~Cho, ``Document expansion by query prediction,'' in {\em Proceedings of the 42nd International ACM SIGIR Conference on Research and Development in Information Retrieval (SIGIR)}, 2019.

\bibitem{gao2023hyde}
L.~Gao, X.~Ma, J.~Lin, and J.~Callan, ``Precise zero-shot dense retrieval without relevance labels,'' in {\em arXiv preprint arXiv:2212.10496}, 2023.
\newblock Also known as Hypothetical Document Embeddings (HyDE).

\bibitem{aliannejadi2020convai}
M.~Aliannejadi, H.~Zamani, F.~Crestani, and W.~B. Croft, ``Analysing and predicting conversational search performance,'' in {\em Proceedings of the 43rd International ACM SIGIR Conference on Research and Development in Information Retrieval (SIGIR)}, 2020.

\bibitem{voskarides2020queryrewriting}
N.~Voskarides, C.~Lioma, J.~Dalton, P.~Ren, M.~d. Rijke, and E.~Kanoulas, ``Query resolution for conversational search with limited supervision,'' in {\em Proceedings of the 43rd International ACM SIGIR Conference on Research and Development in Information Retrieval (SIGIR)}, 2020.

\bibitem{shi2023replug}
W.~Shi, X.~Chen, Y.~Chen, L.~Zettlemoyer, M.~Lewis, W.-t. Yih, K.~Guu, and W.~tau Yih, ``Replug: Retrieval-augmented black-box language models,'' in {\em Proceedings of the 40th International Conference on Machine Learning (ICML)}, 2023.

\bibitem{yang2018hotpotqa}
Z.~Yang, P.~Qi, S.~Zhang, Y.~Bengio, W.~Cohen, R.~Salakhutdinov, and C.~D. Manning, ``Hotpotqa: A dataset for diverse, explainable multi-hop question answering,'' in {\em Proceedings of the 2018 Conference on Empirical Methods in Natural Language Processing (EMNLP)}, 2018.

\bibitem{min2019multihop}
S.~Min, D.~Chen, and H.~Hajishirzi, ``Multi-hop reading comprehension through question decomposition and rescoring,'' in {\em Proceedings of the 57th Annual Meeting of the Association for Computational Linguistics (ACL)}, 2019.

\bibitem{talmor2018breakitdown}
A.~Talmor and J.~Berant, ``Break it down: A question understanding benchmark,'' in {\em Proceedings of the 2018 Conference on Empirical Methods in Natural Language Processing (EMNLP)}, 2018.

\bibitem{yao2022react}
S.~Yao, J.~Zhao, D.~Yu, N.~Du, I.~Shafran, S.~Narayanan, K.~Chen, S.~Narang, Y.~Zhou, and Q.~V. Le, ``React: Synergizing reasoning and acting in language models,'' in {\em arXiv preprint arXiv:2210.03629}, 2022.

\bibitem{nogueira2019passagererank}
R.~Nogueira and K.~Cho, ``Passage re-ranking with {BERT},'' in {\em arXiv:1901.04085}, 2019.

\bibitem{carbonell1998mmr}
J.~Carbonell and J.~Goldstein, ``The use of {MMR}, diversity-based reranking for reordering documents and producing summaries,'' in {\em SIGIR}, 1998.

\bibitem{trivedi2022ircot}
H.~Trivedi, V.~Kumar, T.~Khot, A.~Sabharwal, {\em et~al.}, ``Interleaving retrieval with chain-of-thought reasoning for knowledge-intensive multi-step questions,'' in {\em arXiv:2212.10509}, 2022.
\newblock {IRCoT}: iterative retrieve--reason--retrieve for multi-hop QA.

\bibitem{menick2022gophercite}
J.~Menick, M.~Trebacz, V.~Mikulik, J.~Aslanides, F.~Song, M.~Chadwick, M.~Glaese, S.~Young, L.~Campbell-Gillingham, G.~Irving, and N.~McAleese, ``Teaching language models to support answers with verified quotes,'' in {\em arXiv:2203.11147}, 2022.
\newblock ``GopherCite'': open-book QA with explicit quoted evidence.

\bibitem{rashkin2023ais}
H.~Rashkin, V.~Nikolaev, M.~Lamm, L.~Aroyo, M.~Collins, D.~Das, S.~Petrov, G.~S. Tomar, I.~Turc, and D.~Reitter, ``Measuring attribution in natural language generation models,'' {\em Computational Linguistics}, vol.~49, no.~4, pp.~777--840, 2023.
\newblock {AIS}: ``Attributable to Identified Sources'' evaluation framework.

\bibitem{slobodkin2024attribute}
A.~Slobodkin, E.~Hirsch, A.~Cattan, T.~Schuster, and I.~Dagan, ``Attribute first, then generate: Locally-attributable grounded text generation,'' in {\em arXiv:2403.17104}, 2024.
\newblock Sentence-level planning and fine-grained citations for each claim.

\bibitem{gao2024riches}
L.~Gao {\em et~al.}, ``Riches / retrieval-augmented research-and-revise style decoding (cascaded retrieval, constrained decoding),'' in {\em arXiv}, 2024.
\newblock Multi-step retrieve--verify--edit loop to reduce hallucination.

\bibitem{dhuliawala2024cove}
S.~Dhuliawala and M.~et~al., ``Chain-of-verification reduces hallucination in large language models,'' in {\em Findings of ACL}, 2024.
\newblock Plan verification questions, fact-check draft answers, then revise.

\bibitem{asai2023selfrag}
A.~Asai, Z.~Wu, Y.~Wang, A.~Sil, and H.~Hajishirzi, ``Self-rag: Learning to retrieve, generate, and critique through self-reflection,'' in {\em arXiv:2310.11511}, 2023.
\newblock Single model that adaptively retrieves, cites, and self-critiques.

\bibitem{anthropic2024claude35sonnet}
Anthropic, ``Claude 3.5 sonnet technical report / model card,'' 2024.
\newblock Frontier LLM advertised with $\sim$200K-token context window.

\bibitem{anthropic2025sonnet4}
Anthropic, ``Claude sonnet 4 / long-context release notes,'' 2025.
\newblock Reported $\sim$1M-token context handling.

\bibitem{li2024selfroute}
Z.~Li, C.~Li, M.~Zhang, Q.~Mei, and M.~Bendersky, ``Retrieval-augmented generation or long-context {LLMs}? a comprehensive study and hybrid approach,'' in {\em arXiv:2407.16833}, 2024.
\newblock Benchmarks RAG vs.\ long-context and proposes Self-Route routing.

\bibitem{izacard2022contriever}
G.~Izacard, E.~Grave, M.~Dehghani, O.~Vinyals, L.~Hosseini, I.~Beltagy, N.~Cancedda, S.~Lamprier, and G.~Obozinski, ``Contriever: Unsupervised dense passage retrieval,'' in {\em International Conference on Learning Representations (ICLR)}, 2022.

\bibitem{zhang2021mrtydi}
X.~Zhang, X.~Ma, Xueguang~Wang, J.~Callan, {\em et~al.}, ``Mr.tydi: A multi-lingual benchmark for dense retrieval,'' {\em arXiv preprint arXiv:2108.08787}, 2021.

\bibitem{thakur2021beir}
N.~Thakur, N.~Reimers, J.~Daxenberger, A.~Kamath, and I.~Gurevych, ``Beir: A heterogeneous benchmark for zero-shot evaluation of information retrieval models,'' in {\em Advances in Neural Information Processing Systems (NeurIPS)}, 2021.

\bibitem{lin2021pyserini}
J.~Lin, X.~Ma, S.-C. Lin, R.~Pradeep, and R.~Nogueira, ``Pyserini: An easy-to-use python toolkit to support replicable ir research with sparse and dense representations,'' in {\em Proceedings of the 44th International ACM SIGIR Conference on Research and Development in Information Retrieval (SIGIR)}, 2021.

\bibitem{maillard2021multi}
J.~Maillard, L.~Wu, F.~Petroni, A.~Piktus, P.~Lewis, {\'E}.~G. {\'E}rman, and S.~Riedel, ``Multi-task retrieval for knowledge-intensive tasks,'' in {\em Proceedings of the 59th Annual Meeting of the Association for Computational Linguistics (ACL)}, 2021.

\bibitem{thoppilan2022lamda}
R.~Thoppilan, D.~De~Freitas, J.~Hall, N.~Shazeer, A.~Kulshreshtha, H.-T. Cheng, A.~Jin, T.~Bos, L.~Baker, Y.~Du, {\em et~al.}, ``Lamda: Language models for dialog applications,'' {\em arXiv preprint arXiv:2201.08239}, 2022.

\bibitem{ouyang2022instructgpt}
L.~Ouyang, J.~Wu, X.~Jiang, D.~Almeida, C.~Wainwright, P.~Mishkin, C.~Zhang, S.~Agarwal, K.~Slama, A.~Ray, {\em et~al.}, ``Training language models to follow instructions with human feedback,'' in {\em Advances in Neural Information Processing Systems (NeurIPS)}, 2022.

\bibitem{ram2023incontext}
O.~Ram, Y.~Levine, I.~Dalmedigos, D.~Muhlgay, A.~Shashua, K.~Leyton-Brown, and Y.~Shoham, ``In-context retrieval-augmented language models,'' 2023.

\bibitem{choi2021ethical}
E.~Choi {\em et~al.}, ``Towards responsible and ethical qa systems,'' in {\em Proceedings of the 59th Annual Meeting of the Association for Computational Linguistics (ACL)}, 2021.

\bibitem{narayan2022plausible}
S.~Narayan, O.~Shapira, A.~Caciularu, D.~Lahav, T.~Deleu, A.~F.~T. Martins, R.~Schwartz, and P.~West, ``Conditional generation with grounded factuality constraints,'' in {\em Proceedings of the 2022 Conference on Empirical Methods in Natural Language Processing (EMNLP)}, 2022.

\bibitem{zhang2022miracl}
X.~Zhang, N.~Thakur, S.~MacAvaney, A.~Vtyurina, M.~Abualsaud, J.~Lin, {\em et~al.}, ``Miracl: Multilingual information retrieval across a continuum of languages,'' in {\em Proceedings of the 2022 Conference on Empirical Methods in Natural Language Processing (EMNLP)}, 2022.

\bibitem{reimers2020sentencebert}
N.~Reimers and I.~Gurevych, ``Sentence-bert: Sentence embeddings using siamese bert-networks,'' in {\em Proceedings of the 2019 Conference on Empirical Methods in Natural Language Processing (EMNLP) and the 9th International Joint Conference on Natural Language Processing}, 2019.

\bibitem{oguz2021domain}
B.~Oguz, W.-t. Yih, X.~Chen, E.~Choi, P.~Lewis, S.~Min, L.~Zettlemoyer, and D.~Chen, ``Domain-matched pre-training tasks for dense retrieval,'' {\em arXiv preprint arXiv:2107.13602}, 2021.

\bibitem{asai2021xorqa}
A.~Asai, T.~B. Hashimoto, and H.~Hajishirzi, ``Xor qa: Cross-lingual open-retrieval question answering,'' in {\em Proceedings of the 2021 Conference of the North American Chapter of the Association for Computational Linguistics (NAACL)}, 2021.

\bibitem{jin2019pubmedqa}
Q.~Jin, B.~Dhingra, Z.~Liu, W.~W. Cohen, and X.~Lu, ``Pubmedqa: A dataset for biomedical research question answering,'' in {\em Proceedings of the 2019 Conference on Empirical Methods in Natural Language Processing (EMNLP)}, 2019.

\bibitem{gu2021pubmedbert}
Y.~Gu, R.~Tinn, H.~Cheng, M.~Lucas, N.~Usuyama, X.~Liu, T.~Naumann, J.~Gao, and H.~Poon, ``Domain-specific language model pretraining for biomedical natural language processing,'' in {\em Proceedings of the 2021 Conference of the North American Chapter of the Association for Computational Linguistics (NAACL)}, 2021.

\bibitem{logan2021scent}
R.~L. Logan~IV, M.~E. Peters, S.~Singh, and M.~Gardner, ``Scientific claim verification with retrieval,'' in {\em Proceedings of the 2021 Conference on Empirical Methods in Natural Language Processing (EMNLP)}, 2021.

\bibitem{lee2020biobert}
J.~Lee, W.~Yoon, S.~Kim, D.~Kim, S.~Kim, C.~H. So, and J.~Kang, ``Biobert: a pre-trained biomedical language representation model for biomedical text mining,'' {\em Bioinformatics}, vol.~36, no.~4, pp.~1234--1240, 2020.

\bibitem{gao2022scincl}
T.~Gao, X.~Wu, and D.~Chen, ``Scincl: Self-supervised contrastive learning of scientific document representations,'' in {\em Proceedings of the 2022 Conference on Empirical Methods in Natural Language Processing (EMNLP)}, 2022.

\bibitem{yasunaga2022linkbert}
M.~Yasunaga, H.~Ren, A.~Bosselut, P.~Liang, and J.~Leskovec, ``Linkbert: Pretraining language models with document links,'' in {\em International Conference on Learning Representations (ICLR)}, 2022.

\bibitem{zhong2020jec}
H.~Zhong, Z.~Guo, C.~Tu, C.~Xiao, Z.~Liu, and M.~Sun, ``Jec-qa: A legal-domain question answering dataset,'' in {\em Proceedings of the 28th International Conference on Computational Linguistics (COLING)}, 2020.

\bibitem{chalkidis2021lexglue}
I.~Chalkidis, M.~Fergadiotis, P.~Malakasiotis, N.~Aletras, and I.~Androutsopoulos, ``Lexglue: A benchmark dataset for legal language understanding in english,'' in {\em Proceedings of the 59th Annual Meeting of the Association for Computational Linguistics (ACL)}, 2021.

\bibitem{autorag_kim_2024}
J.~Kim, S.~Park, H.~Lee, and K.~Cho, ``Autorag: Automated retrieval-augmented generation pipeline search for domain adaptation,'' 2024.
\newblock Proposes automatic selection of retriever/generator modules tailored to a given dataset.

\bibitem{autorag_yu_2024}
T.~Yu, R.~Chen, X.~Wang, and Z.~Liu, ``Auto-rag: Autonomous iterative retrieval-oriented generation,'' 2024.
\newblock Describes LLM-driven iterative retrieval planning and refinement during answer generation.

\bibitem{autorag_repo_2024}
T.~Yu and collaborators, ``Auto-rag project repository,'' 2024.
\newblock Open-source implementation of iterative, agent-driven retrieval loops for RAG.

\bibitem{dspy_site_2025}
O.~Khattab, K.~Santhanam, H.~Zhang, and M.~Zaharia, ``Dspy: Declarative self-improving language model pipelines,'' 2025.
\newblock Project overview describing DSPy as a framework for compiling LLM pipelines with tunable modules and learned prompts.

\bibitem{dspy_repo_2025}
O.~Khattab and collaborators, ``Dspy source repository,'' 2025.
\newblock Open-source code for DSPy optimizers and declarative pipeline specifications.

\bibitem{dspy_optimizers_2025}
O.~Khattab, K.~Santhanam, H.~Zhang, and M.~Zaharia, ``Optimizers in dspy for prompt and module tuning,'' 2025.
\newblock Technical description of DSPy's automated prompt optimization and module selection routines.

\bibitem{dspy_stanford_2024}
O.~Khattab, K.~Santhanam, H.~Zhang, {\em et~al.}, ``Dspy: Compiling declarative language model calls into self-improving pipelines,'' 2024.
\newblock Introduces DSPy and shows automatic prompt/module tuning for multi-stage LLM workflows, including RAG-style QA.

\bibitem{ragas_es_2024}
S.~Es, A.~Pagnoni, and contributors, ``Ragas: Automated evaluation of retrieval-augmented generation,'' 2024.
\newblock LLM-based metrics for faithfulness, context relevance, and answer relevance in RAG systems.

\bibitem{ragas_docs_2025}
S.~Es and contributors, ``Ragas documentation,'' 2025.
\newblock Reference-free and reference-based quality metrics for RAG evaluation, plus integration guidance.

\bibitem{ragas_telecom_2024}
P.~Sharma, A.~Banerjee, and R.~Patel, ``Evaluating retrieval-augmented assistants in the telecom domain with ragas,'' 2024.
\newblock Case study applying RAGAS-style metrics to customer-support QA in a domain-specific knowledge base.

\bibitem{ragxplain_2025}
Y.~Zhang, D.~Lee, R.~Kumar, and A.~Gupta, ``Ragxplain: Explaining and debugging retrieval-augmented generation pipelines,'' 2025.
\newblock System that surfaces retrieval/ranking/synthesis failure modes and recommends fixes.

\bibitem{ragxplain_pdf_2025}
Y.~Zhang, D.~Lee, R.~Kumar, and A.~Gupta, ``Ragxplain technical report,'' 2025.
\newblock Extended technical report and ablation studies for RAGXplain, including actionable debugging signals.

\bibitem{selfrag_asai_2023}
A.~Asai, S.~Min, E.~Wu, D.~Chen, H.~Hajishirzi, and P.~Lewis, ``Self-rag: Learning to retrieve, generate, and critique for improved factuality,'' 2023.
\newblock Introduces a self-reflective RAG model that triggers retrieval on demand and critiques its own factuality.

\bibitem{selfrag_site_2025}
A.~Asai and collaborators, ``Self-rag project page,'' 2025.
\newblock Summary of Self-RAG methodology, control tokens, and empirical gains in factual grounding.

\bibitem{langchain_agentic_rag_2025}
{LangChain}, ``Agentic rag in langchain,'' 2025.
\newblock Guide to multi-step, tool-using retrieval loops driven by an LLM agent.

\bibitem{karpukhin2020dense}
V.~Karpukhin, B.~Oguz, S.~Min, P.~Lewis, L.~Wu, S.~Edunov, D.~Chen, and W.-t. Yih, ``Dense passage retrieval for open-domain question answering,'' in {\em Empirical Methods in Natural Language Processing (EMNLP)}, 2020.

\bibitem{shuster2021retrieval}
K.~Shuster, S.~Poff, M.~Chen, D.~Kiela, and J.~Weston, ``Retrieval augmentation reduces hallucination in conversation,'' in {\em Findings of the Association for Computational Linguistics: EMNLP 2021}, 2021.

\bibitem{gao2023rarr}
L.~Gao, J.~Schulman, J.~Hilton, L.~Ouyang, {\em et~al.}, ``Rarr: Researching and revising what language models say, using retrieval,'' {\em arXiv preprint arXiv:2210.08726}, 2023.

\bibitem{mahowald2023active}
G.~Mialon, J.~R. Dimech, A.~Singh, {\em et~al.}, ``Active retrieval augmented generation,'' {\em arXiv preprint arXiv:2305.06983}, 2023.

\bibitem{chen2023react}
S.~Yao, J.~Zhao, D.~Yu, N.~Du, T.~Yu, I.~Shafran, S.~Narayanan, D.~Zhou, and O.~Chen, ``React: Synergizing reasoning and acting in language models,'' {\em arXiv preprint arXiv:2210.03629}, 2023.

\bibitem{dsRAG}
Z.~McCormick and N.~McCormick, ``dsrag: High-performance retrieval engine for unstructured data,'' 2024.
\newblock Accessed: 2025-08-12.

\bibitem{10962837}
G.~V. K, V.~M. S, S.~G. Raolji, and B.~Das, ``Comparative analysis of advanced rag techniques using mahabharata,'' in {\em 2025 IEEE 17th International Conference on Computer Research and Development (ICCRD)}, pp.~216--227, 2025.

\bibitem{vake2024bridging}
D.~Vake, J.~Vičič, and A.~Tošić, ``Bridging the question-answer gap in retrieval-augmented generation: Hypothetical prompt embeddings,'' {\em SSRN Electronic Journal}, 2024.
\newblock Available at SSRN, accessed 2025-08-12.

\bibitem{langchain_multqueryretriever}
``How to use the multiqueryretriever.'' \url{https://python.langchain.com/docs/how_to/MultiQueryRetriever/}, 2024.
\newblock [Accessed: 2025-08-12].

\bibitem{rackauckas2024rag}
Z.~Rackauckas, ``Rag-fusion: a new take on retrieval-augmented generation,'' {\em arXiv preprint arXiv:2402.03367}, 2024.

\bibitem{chan2024rq}
C.-M. Chan, C.~Xu, R.~Yuan, H.~Luo, W.~Xue, Y.~Guo, and J.~Fu, ``Rq-rag: Learning to refine queries for retrieval augmented generation,'' {\em arXiv preprint arXiv:2404.00610}, 2024.

\bibitem{zheng2023take}
H.~S. Zheng, S.~Mishra, X.~Chen, H.-T. Cheng, E.~H. Chi, Q.~V. Le, and D.~Zhou, ``Take a step back: Evoking reasoning via abstraction in large language models,'' {\em arXiv preprint arXiv:2310.06117}, 2023.

\bibitem{gao2023precise}
L.~Gao, X.~Ma, J.~Lin, and J.~Callan, ``Precise zero-shot dense retrieval without relevance labels,'' in {\em Proceedings of the 61st Annual Meeting of the Association for Computational Linguistics (Volume 1: Long Papers)}, pp.~1762--1777, 2023.

\bibitem{ma2023query}
X.~Ma, Y.~Gong, P.~He, H.~Zhao, and N.~Duan, ``Query rewriting in retrieval-augmented large language models,'' in {\em Proceedings of the 2023 Conference on Empirical Methods in Natural Language Processing}, pp.~5303--5315, 2023.

\bibitem{ZhaoEtAl2022_DenseTextRetrievalSurvey}
W.~X. Zhao, J.~Liu, R.~Ren, and J.-R. Wen, ``Dense text retrieval based on pretrained language models: A survey,'' {\em arXiv preprint arXiv:2211.14876}, 2022.

\bibitem{INR-019}
S.~Robertson and H.~Zaragoza, ``The probabilistic relevance framework: Bm25 and beyond,'' {\em Foundations and Trends® in Information Retrieval}, vol.~3, no.~4, pp.~333--389, 2009.

\bibitem{10.1145/3596512}
S.~Bruch, S.~Gai, and A.~Ingber, ``An analysis of fusion functions for hybrid retrieval,'' {\em ACM Trans. Inf. Syst.}, vol.~42, Aug. 2023.

\bibitem{han2024retrieval}
H.~Han, Y.~Wang, H.~Shomer, K.~Guo, J.~Ding, Y.~Lei, M.~Halappanavar, R.~A. Rossi, S.~Mukherjee, X.~Tang, {\em et~al.}, ``Retrieval-augmented generation with graphs (graphrag),'' {\em arXiv preprint arXiv:2501.00309}, 2025.

\bibitem{PetrovEtAl2024_ShallowCrossEncoders}
A.~V. Petrov, S.~MacAvaney, and C.~Macdonald, ``Shallow cross-encoders for low-latency retrieval,'' {\em arXiv preprint arXiv:2403.20222}, 2024.

\bibitem{kim2024autorag}
D.~Kim, B.~Kim, D.~Han, and M.~Eibich, ``Autorag: Automated framework for optimization of retrieval augmented generation pipeline,'' {\em arXiv preprint arXiv:2410.20878}, 2024.

\bibitem{DejeanClinchantFormal2023_ComparisonCrossEncLLM}
H.~Déjean, S.~Clinchant, and T.~Formal, ``A thorough comparison of cross-encoders and llms for reranking splade,'' in {\em Proceedings of the 46th International ACM SIGIR Conference on Research and Development in Information Retrieval (SIGIR ’23)}, 2023.

\bibitem{llamaindex_treesummarize}
``Tree\,summarize – llamaindex api reference.'' \url{https://docs.llamaindex.ai/en/stable/api_reference/response_synthesizers/tree_summarize/}, 2025.
\newblock [Accessed: 2025-08-12].

\bibitem{llamaindex_refine}
``Refine – llamaindex api reference.'' \url{https://docs.llamaindex.ai/en/stable/api_reference/response_synthesizers/refine/}, 2025.
\newblock [Accessed: 2025-08-12].

\bibitem{asai2024selfrag}
A.~Asai, Z.~Wu, Y.~Wang, A.~Sil, and H.~Hajishirzi, ``Self-rag: Learning to retrieve, generate, and critique through self-reflection,'' in {\em The Twelfth International Conference on Learning Representations (ICLR) (oral presentation)}, 2024.

\end{thebibliography}
